\documentclass[10pt,twocolumn,letterpaper]{article}

\usepackage{cvpr}              

\usepackage{graphicx, caption, subcaption, multirow, overpic, textpos}
\usepackage[table]{xcolor}

\newlength\savewidth\newcommand\shline{\noalign{\global\savewidth\arrayrulewidth
  \global\arrayrulewidth 1pt}\hline\noalign{\global\arrayrulewidth\savewidth}}
\newcommand{\tablestyle}[2]{\setlength{\tabcolsep}{#1}\renewcommand{\arraystretch}{#2}\centering\footnotesize}
\renewcommand{\paragraph}[1]{\vspace{0.1mm}\noindent\textbf{#1}}

\newcolumntype{x}[1]{>{\centering\arraybackslash}p{#1pt}}
\newcolumntype{y}[1]{>{\raggedright\arraybackslash}p{#1pt}}
\newcolumntype{z}[1]{>{\raggedleft\arraybackslash}p{#1pt}}

\newcommand{\app}{\raise.17ex\hbox{$\scriptstyle\sim$}}

\definecolor{deemph}{gray}{0.6}

\definecolor{baselinecolor}{gray}{.9}
\newcommand{\baseline}[1]{\cellcolor{baselinecolor}{#1}}

\usepackage{transparent}
\usepackage{algorithm}
\usepackage{algorithmic}
\usepackage{amsmath}
\usepackage{amssymb}
\usepackage{mathtools}
\usepackage{booktabs}
\usepackage{tabularx}
\usepackage[export]{adjustbox}
\usepackage{multirow}
\usepackage{colortbl}
\usepackage{makecell}
\usepackage{bbding}
\usepackage{amssymb}
\usepackage{pifont}

\usepackage{textcomp}

\newcommand{\xd}[1]{{\color{orange} #1}}

\def\ie{\textit{i.e.}}
\def\eg{\textit{e.g.}}

\definecolor{cvprblue}{rgb}{0.21,0.49,0.74}
\usepackage[pagebackref,breaklinks,colorlinks,citecolor=cvprblue]{hyperref}

\def\paperID{6152} 
\def\confName{CVPR}
\def\confYear{2024}
\title{Holistic Autonomous Driving Understanding by Bird's-Eye-View Injected \\ Multi-Modal Large Models}

\author{Xinpeng Ding\textsuperscript{1}
\quad
Jianhua Han\textsuperscript{2}
\quad 
Hang Xu\textsuperscript{2}
\quad
Xiaodan Liang\textsuperscript{3}
\quad
Wei Zhang\textsuperscript{2}
\quad
Xiaomeng Li\textsuperscript{1}
\vspace{0.4em} 
\\
\textsuperscript{1} Hong Kong University of Science and Technology
\quad
\textsuperscript{2}Huawei Noah’s Ark Lab
\quad \\
\textsuperscript{3}Sun Yat-Sen University
\\
\vspace{-1mm}
\url{https://github.com/xmed-lab/NuInstruct}
}

\begin{document}
\newcommand{\dataset}{NuInstruct}
\newcommand{\AD}{Autonomous Driving}
\newcommand{\ad}{autonomous driving}
\definecolor{deepgreen}{rgb}{0.0, 0.6, 0.0}
\newcommand{\cmark}{\scalebox{0.9}{\textcolor{deepgreen}{\ding{52}}}}
\newcommand{\xmark}{\textcolor{red}{{\ding{55}}}}
\definecolor{mygray}{gray}{.9}
\definecolor{cream}{rgb}{1.0, 0.99, 0.82}
\definecolor{aliceblue}{rgb}{0.94, 0.97, 1.0}
\definecolor{beaublue}{rgb}{0.74, 0.83, 0.9}
\definecolor{non-photoblue}{rgb}{0.64, 0.87, 0.93}

\definecolor{myyellow}{rgb}{0.71, 0.55, 0.0}
\definecolor{myblue}{rgb}{0.0, 0.71, 0.71}
\definecolor{color2}{rgb}{0.55, 0.71, 0.0}
\definecolor{color1}{rgb}{0.98, 0.81, 0.69}
\definecolor{color3}{rgb}{1.0, 0.6, 0.4}
\definecolor{color4}{rgb}{0.29, 0.59, 0.82}
\definecolor{myred}{rgb}{0.81, 0.0, 0.0}
\definecolor{myblue}{rgb}{0.0, 0.71, 0.71}
\definecolor{mygreen}{rgb}{0.0, 0.51, 0.0}
\newcommand{\defaultsetting}[1]{\colorbox{lightgreen}{#1}}
\twocolumn[{%
\maketitle
\begin{center}
    \centering
    \includegraphics[width=\textwidth]{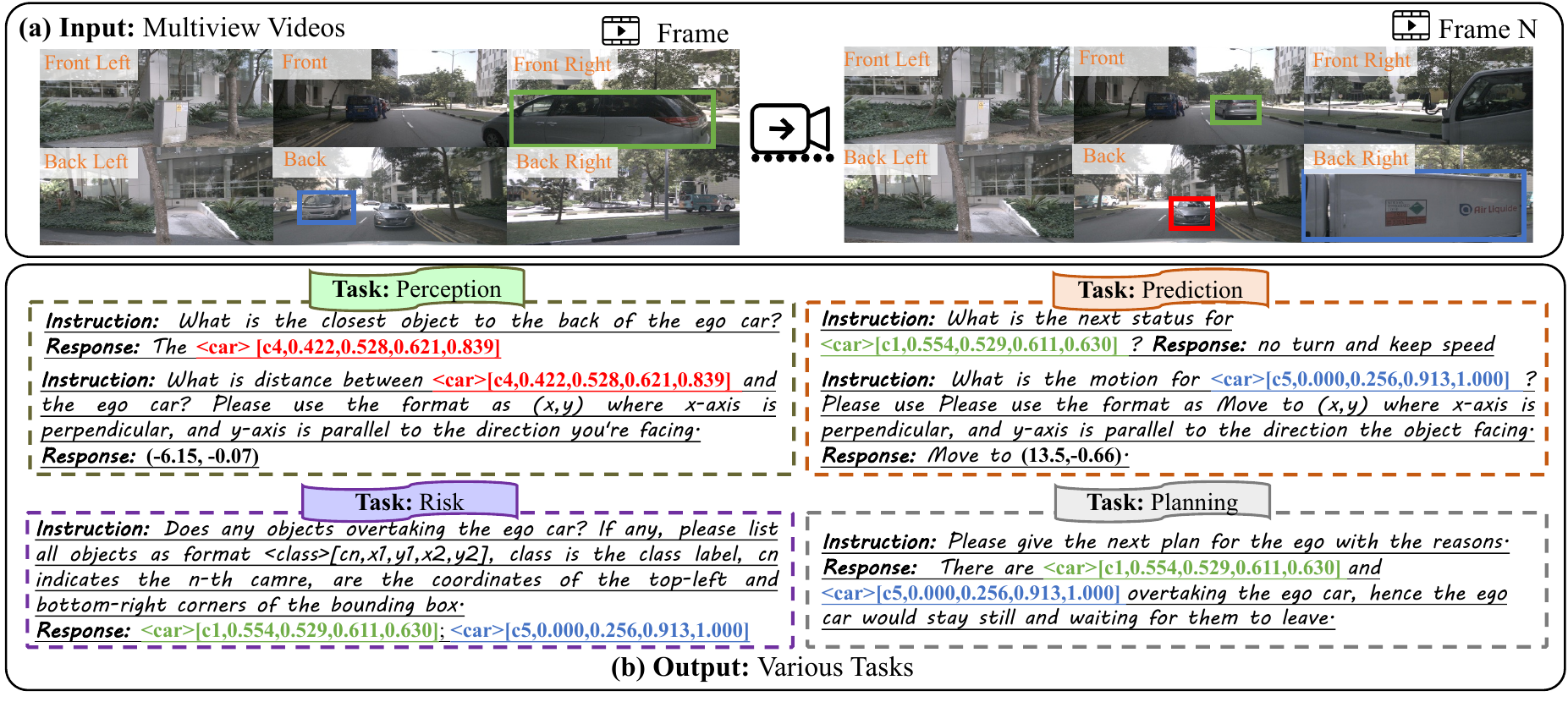}
    \vspace{-6mm}
    \captionof{figure}[Teaser]{\textbf{Example of our proposed \dataset~dataset for holistic language-based \ad.} \textbf{(a)} The input are multi-view videos.
    \textbf{(b)} Various tasks are presented in instruction-response format. There are a total of four tasks, covering $17$ subtasks (see in Fig.\ref{fig:statistics}~(a)). 
    }

    \label{fig:teaser}
\end{center}%
}]
\begin{abstract}
The rise of multimodal large language models (MLLMs) has spurred interest in language-based driving tasks. However, existing research typically focuses on limited tasks and often omits key multi-view and temporal information which is crucial for robust autonomous driving. To bridge these gaps, we introduce \dataset, a novel dataset with 91K multi-view video-QA pairs across 17 subtasks, where each task demands holistic information (~\eg, temporal, multi-view, and spatial), significantly elevating the challenge level. To obtain \dataset, we propose a novel SQL-based method to generate instruction-response pairs automatically, which is inspired by the driving logical progression of humans. We further present BEV-InMLLM, an end-to-end method for efficiently deriving instruction-aware Bird’s-Eye-View (BEV) features, language-aligned for large language models. BEV-InMLLM integrates multi-view, spatial awareness, and temporal semantics to enhance MLLMs' capabilities on \dataset~tasks. Moreover, our proposed BEV injection module is a plug-and-play method for existing MLLMs. Our experiments on \dataset~demonstrate that BEV-InMLLM significantly outperforms existing MLLMs,~\eg ~$9\%$ improvement on various tasks. We plan to release our \dataset~for future research development.
%
%
%

\if
Multimodal large language models (MLLMs) have demonstrated impressive\xd{some people may disagree the performance is good, just recently explored, can replace with a more moderate word} performance in autonomous driving tasks, such as perception, planning, and interpretability. Most existing benchmarks and methods focus on understanding from a single front-view perspective. However, multi-view (MV) information is very critical for driving scenarios,~\eg, offering the surrounding scene of the ego car, thus leading to more reliable and safe decision-making.
In this paper, we propose to inject MV information into MLLMs and introduce the MVGPT.
Specifically, MVGPT can take MV videos as input, and perform diverse multi-view tasks for autonomous driving, including, perception, prediction, reasoning, and planning.
To this end, we first introduce MultiviewQA, a multi-view video-language autonomous driving dataset containing 110K question-answer pairs.\xd{move after finishing the MVGPT part}
%
To efficiently train MVGPT, we utilize existing single-view MLLMs as our frozen backbones, and only train a lightweight BEV adapter to enable them for MV video-language understanding.
%
The proposed BEV adapter incorporates the high-resolution, location-aware bird’s-eye-view (BEV) representations into MLLMs to better capture the surrounding scenes, thus improving the MV understanding. 
%
%
Extensive experiments on MultiviewQA show that our approach significantly outperforms existing single-view MLLMs,~\eg, xxxx.
\fi

\end{abstract}
\begin{table*}
\centering
\begin{adjustbox}{max width=\linewidth}
\begin{tabular}{l|cccc|cccccc|c}
\hline
\rowcolor{mygray} & \multicolumn{4}{c|}{\it \textbf{Tasks}}  & \multicolumn{6}{c|}{\it \textbf{Information}} & \\
\rowcolor{mygray}\multirow{-2}{*}{ \it \textbf{Dataset}}& {\it \small Perception} & {\it \small Prediction} & {\it \small Risk} &{\it \small P w/ R}  &   {\it \small Multi-view} & {\it \small Temporal} & {\it \small Multi-object} & {\it \small Distance} & {\it \small Position} & {\it \small Road} &  {\multirow{-2}{*}{\it \textbf{Scale}}} \\
\hline
\hline
BDD-X~\cite{kim2018textual}  & \xmark & \xmark & \xmark & \cmark &  \xmark &  \cmark &  \xmark & \xmark & \xmark & \xmark & $20$K \\
Talk2Car~\cite{deruyttere2019talk2car} & \cmark & \xmark & \xmark & \xmark &  \xmark &  \cmark &  \xmark & \xmark & \xmark & \xmark & $11$K\\
DRAMA~\cite{malla2023drama} & \cmark & \xmark& \cmark & \xmark &  \xmark &  \cmark &  \xmark & \xmark &  \xmark & \xmark  & $100$K \\
DRAMA\small{-ROLISP}~\cite{ding2023hilm} & \cmark & \xmark& \cmark & \cmark &  \xmark &  \cmark &  \xmark & \xmark &  \xmark & \xmark & $35$K \\
DriveGPT4~\cite{xu2023drivegpt4} & \cmark & \xmark & \cmark & \xmark &  \xmark &  \cmark &  \cmark & \xmark & \xmark & \cmark& $28$K \\
Talk2BEV~\cite{dewangan2023talk2bev} & \cmark & \cmark & \xmark & \cmark &  \xmark &  \xmark &  \cmark & \cmark & \cmark& \xmark & $20$K  \\
Nuscenes-QA~\cite{qian2023nuscenes} &  \cmark & \xmark& \xmark & \xmark &  \cmark &  \xmark &  \cmark & \cmark &  \cmark  & \xmark & $459$K \\
NuPrompt~\cite{wu2023language} & \cmark & \xmark& \xmark & \xmark &  \cmark &  \cmark &  \cmark & \xmark &  \xmark  &  \xmark& $35$K  \\
\hline
NuInstruct (Ours)  & \cmark & \cmark& \cmark & \cmark &  \cmark &  \cmark &  \cmark & \cmark &  \cmark & \cmark  &  $91$K\\
\hline
\end{tabular}
\end{adjustbox}
\vspace{-0.8em}
\caption{\textbf{Comparison of our NuInstruct with existing language-based driving datasets.} `\emph{P w/ R}' indicates the planning with reasoning.
NuInstruct provides various tasks and comprehensive information (~\eg, including multi-view, temporal, distance, and so on) for comprehensive \ad~ understanding.}
\label{tab:comparison}
\end{table*}

\section{Introduction}

Witnessing the success of multimodal large language models (MLLMs)~\cite{zhu2023minigpt, Dai2023instruct,chen2023shikra,liu2023visual,pi2023detgpt,zhang2023video,li2023videochat,hong20233d}, language-based driving is one of the trends in various autonomous driving tasks~\cite{malla2023drama,ding2023hilm,xu2023drivegpt4,dewangan2023talk2bev}.
For instance, some researchers ground the instruction prompts to single or multiple objects for 2D or 3D object detection and tracking~\cite{deruyttere2019talk2car, Vasudevan_Dai_Gool_2018,wu2023referring,wu2023language,zou2023unim}.
Nuscenes-QA~\cite{qian2023nuscenes} offers numerous question-answer pairs for multi-view perception tasks in driving scenes.
Some advancements,~\eg, DRAMA~\cite{malla2023drama} and HiLM-D~\cite{ding2023hilm}, generating text descriptions for localizing risk objects.
Except for perception tasks, DriveGPT4~\cite{xu2023drivegpt4} and GPT-Driver~\cite{mao2023gpt} leverage LLMs for interpreting vehicle actions and planning, respectively.
Talk2BEV~\cite{dewangan2023talk2bev} formulate BEV into a JSON file and input it into ChatGPT~\cite{OpenAI_2023} to conduct \ad~understanding.

\vspace{-0.5mm}
Although remarkable progress has been achieved, current language-based driving research still exhibits two main shortcomings as shown in Table~\ref{tab:comparison}.
\emph{\textbf{(i) Partial tasks}}. Existing benchmarks only cover a subset of \ad~tasks.
However, autonomous driving comprises a series of interdependent tasks, each indispensable to the system's overall functionality~\cite{caesar2020nuscenes}.
For instance, it is challenging to make reliable predictions when lacking accurate perception.
\emph{\textbf{(ii) Incomplete information}}.
The information utilized by existing methods for executing these tasks is often incomplete.
Specifically, existing datasets~\cite{ding2023hilm,xu2023drivegpt4} usually consist of single-view-based images, without considering temporal and multi-view information.
However, safe driving decisions require a holistic understanding of the environment,~\eg, only concerning on the front may neglect an overtaking vehicle in the left~\cite{safety2040020}. 

To address the above two problems, we first create \textbf{\dataset}, a comprehensive language-driving dataset with $91$K multi-view video-QA pairs across $17$ subtasks (Fig.~\ref{fig:statistics}). Our dataset presents more complex tasks than existing benchmarks, demanding extensive information like multi-view, temporal, distance and so on, as shown in Fig.~\ref{fig:teaser} and Table~\ref{fig:statistics}.
%
%
%
To obtain \dataset, we introduce a SQL-based method for the automated generation of instruction-response pairs. This method transforms instruction-response creation into a process utilizing structured query languages (SQLs)~\cite{date1989guide} from a database.
\emph{Our rationale for the tasks and their corresponding SQL design follows the logical progression of human drivers: \ding{182} initially observing surrounding objects (\textbf{Perception}), \ding{183} predicting their behavior (\textbf{Prediction}), \ding{184} assessing potential threats such as overtaking vehicles (\textbf{Risk}), and \ding{185} ultimately using the previous information to plan a safe route with justified reasoning (\textbf{Planning with Reasoning})}.
Finally, to ensure the quality of our \dataset, we conduct human or GPT-4~\cite{OpenAI_2023} verification to eliminate the erroneous instruction-response pairs.
Compared with other data generation methods,~\eg, ChatGPT-based~\cite{OpenAI_2023} or human-based~\cite{dewangan2023talk2bev}, this structured design ensures the generation of instruction-response pairs is both reliable and scalable.
%

\vspace{-0.3mm}
To address the challenging tasks of the proposed \dataset, we further extend the current MLLMs to receive more holistic information.
Existing MLLMs are constrained by their design for single-view inputs. 
To overcome this, we provide a Multi-View MLLM (MV-MLLM) with a specialized Multi-view Q-Former capable of processing multi-view video inputs. 
Although MV-MLLM allows for the capture of Multi-view temporal appearance features, they often miss out on critical information (\eg, distance, spatial) as well as suffer from occlusions.
%
%
BEV's feature, a formulation of multi-view inputs, has been widely adopted in traditional autonomous driving models since they can clearly represent object locations and scales (essential for distance/spatial-sensitive tasks)~\cite{huang2021bevdet,chen2022relation,chen2023coda}.
Leveraging this, we integrate BEV into MV-MLLM to create BEV-InMLLM, enhancing perception and decision-making in autonomous driving by capturing a comprehensive information spectrum.
Inspired by this, we integrate BEV into MV-MLLM, obtaining BEV-InMLLM to capture a full spectrum of information for reliable perception and decision-making in autonomous driving.
BEV-InMLLM uses a BEV injection module to effectively obtain BEV features aligned with language features for LLMs.
This approach is more resource-efficient than training a BEV extractor from scratch with visual-language data like CLIP~\cite{radford2021learning,Li_2023_CVPR}.
Notably, our BEV injection module serves as a plug-and-play solution for existing MLLM


%

%
Overall, our contributions can be summarized as follows:
\begin{itemize}
    \item We curate \dataset, a new language-driving dataset with $91$K multi-view video-instruction-response pairs across 17 subtasks, using a novel SQL-based method. \dataset~is currently the most holistic language-driving dataset, to our knowledge. We plan to release our \dataset~for future research development.
    \item We propose BEV-InMLMM to integrate instruction-aware BEV features with existing MLLMs, enhancing them with a full suite of information, including temporal, multi-view, and spatial details. Notably, our BEV injection module serves as a plug-and-play solution for existing MLLM.
    \item Our experiments with \dataset~demonstrate our proposed methods significantly boost MLLM performance in various tasks, notably outperforming state-of-the-art by ~9\% on various tasks. 
    Ablation studies show that MV-MLLM enhances multi-view tasks, and BEV-InMLLM is vital for most tasks, emphasizing the importance of spatial information.
\end{itemize}

\if
However, aligning BEV features with language for LLMs poses a significant challenge,~\eg, training a BEV extractor from scratch with visual-language data like CLIP~\cite{radford2021learning} requires substantial data and computational resources.
%
To effectively inject the language-aligned BEV features, we further propose BEV-InMLLM, the enhanced MV-MLLM by a BEV injection module.
The BEV injection module is composed of an instruction-aware BEV Q-Former that selects BEV features relevant to the given instructions and an injection mechanism that merges these with language-aligned multi-view features.
This integration enables BEV-InMLLM to receive a full spectrum of information for reliable perception and decision-making in autonomous driving, as shown in Fig.~\ref{fig:teaser}. 
\fi

\section{Related Work}

\noindent\textbf{Language-driving datasets and models.}
CityScapes-Ref~\cite{Vasudevan_Dai_Gool_2018}, Talk2Car~\cite{deruyttere2019talk2car} perform language-grounding tasks. ReferKITTI~\cite{wu2023referring} and NuPrompt~\cite{qian2023nuscenes} leverage temporal data for 2D or 3D referring object detection and tracking.
Nuscenes-QA~\cite{qian2023nuscenes} offers numerous question-answer pairs for multi-view perception tasks in driving scenes.
Some advancements,~\eg, DRAMA~\cite{malla2023drama} and HiLM-D~\cite{ding2023hilm}, generating text descriptions for localizing risk objects.
Beyond perception, DriveGPT4~\cite{xu2023drivegpt4} and GPT-Driver~\cite{mao2023gpt} leverage LLMs for interpreting vehicle actions and planning, respectively.
Talk2BEV~\cite{dewangan2023talk2bev} formulate BEV into a JSON file and input it into ChatGPT~\cite{OpenAI_2023} to conduct \ad~understanding.
%
Despite these advancements, a common limitation persists: most datasets and models address only part of the \ad~tasks with incomplete information.
As shown in Table~\ref{tab:comparison} and Fig.~\ref{fig:teaser}, in this paper, we propose a challenging dataset containing various tasks that require holistic information,~\ie, temporal, multi-view, spatial and so on, to address. 

\vspace{1mm}
\noindent\textbf{Multimodal Large Language Models.}
Leveraging the capabilities of pre-trained LLMs like LLaMA~\cite{touvron2023llama} and Vicuna~\cite{chiang2023vicuna}, Multimodal LLMs (MLLMs) are expanding their application spectrum, handling inputs from images~\cite{pi2023detgpt, li2022blip,li2023blip,pi2023perceptiongpt,gao2023gllava,zhu2023minigpt,alayrac2022flamingo,chen2023shikra}, videos~\cite{zhang2023video,li2023videochat,li2023etc}, and 3D data~\cite{hong20233d,huang2023clip2point} to medical data~\cite{li2023llava}. In the domain of autonomous driving, DriveGPT4~\cite{xu2023drivegpt4} and Talk2BEV~\cite{dewangan2023talk2bev} have integrated MLLMs for comprehension. However, these approaches have limitations; DriveGPT4 is confined to single-view inputs, and Talk2BEV lacks temporal dynamics and an end-to-end framework. Addressing these gaps, our BEV-InMLLM model assimilates comprehensive temporal, multi-view, and spatial data, for reliable decisions.


\begin{figure}[t]
    \centering
    \includegraphics[width=\linewidth,height=0.18\textheight]{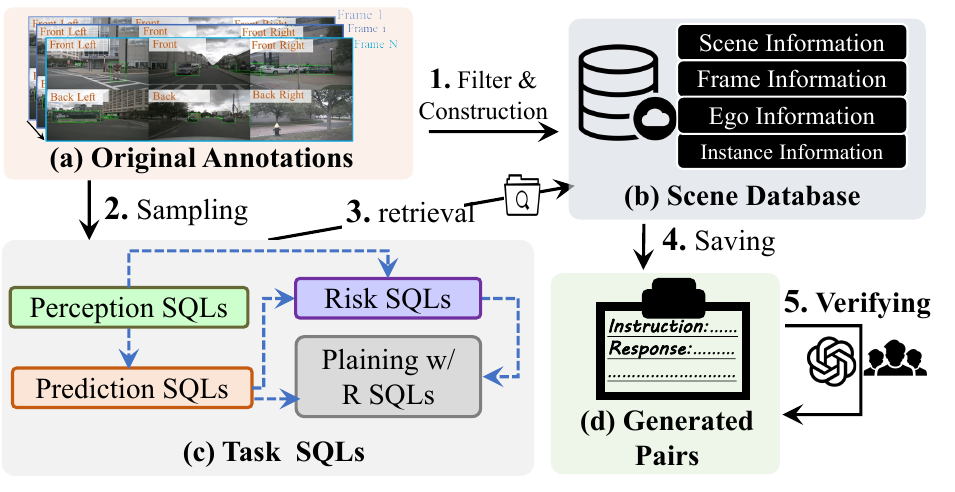}
    \vspace{-2.0em}
    \caption{\textbf{Procedure of SQL-based data generation.} We formulate the data generation into an SQL-based process, using different task SQLs to retrieve the response from the scene information database.
      The design of SQLs follows the logical flow of autonomous driving tasks~\cite{hu2023planning}, which is represented in \textcolor{blue}{blue dashed arrows}.
    `Planning w/ R' indicates the planning with reasoning.
    }
    \label{fig:datageneration}
\end{figure}
\begin{figure*}[t]
    \centering
\includegraphics[width=1.0\textwidth,height=0.22\textheight]{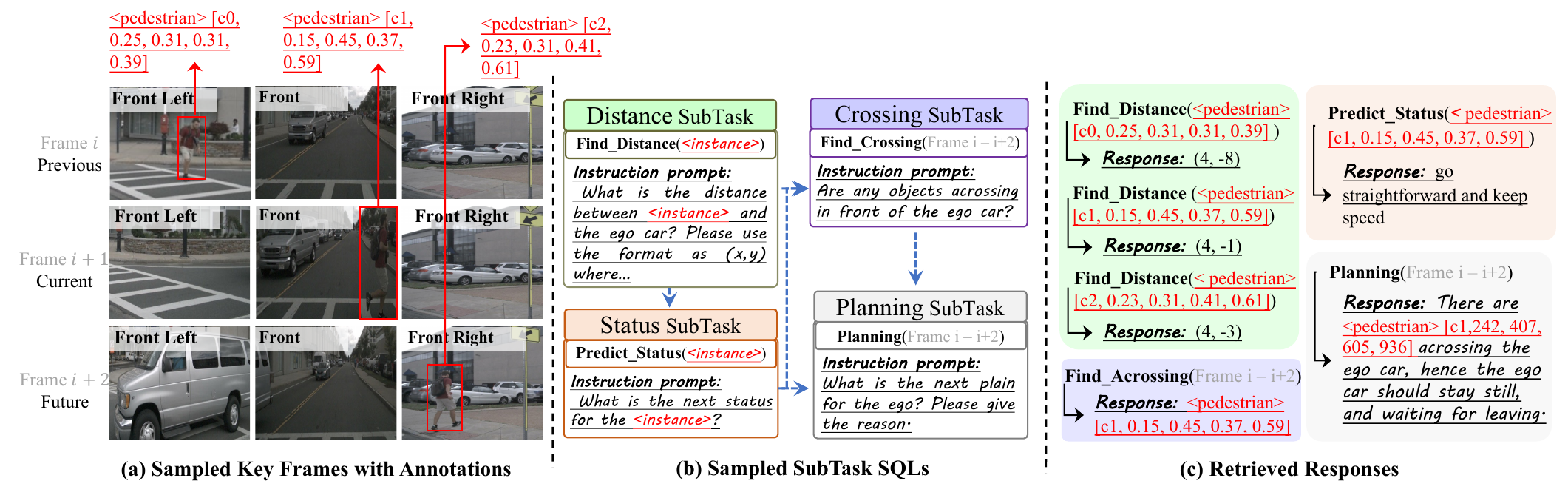}
\vspace{-2.0em}
    \caption{\textbf{The illustration of an example for Step 3 retrieval in the data generation process.} 
    \textbf{(a) Sampled keyframes with annotations.} Three keyframes with annotations are randomly sampled, and we only select one instance,~\ie, the pedestrian (\textcolor{red}{box}), in this example for clarity. 
    %
     \textbf{(b) Sampled subtask SQLs.}
    %
    %
    %
    Each subtask SQL consists of two parts,~\ie, the subtask function and the instruction prompt.
    %
    %
    %
    \textbf{(c) Retrieved Responses.}  The subtask function receives the specific input and retrieves the responses from the scene information database.
    }
    \label{fig:example}
\end{figure*}
\section{NuInstruct}

In this section, we will illustrate the details of our constructed NuInstruct dataset.
In Section~\ref{sec:datagenerate}, we will discuss the process of data generation. We will then go deeper and provide statistics on our dataset in Section~\ref{sec:datastatistics}. Finally, Section~\ref{sec:evaluationmetrics} shows the evaluation metrics to evaluate the performance of different models on the new~\dataset.

\subsection{Instruction Data Generation}~\label{sec:datagenerate}
Our NuInstruct is built on one of the most popular datasets,~\ie, Nuscenes~\cite{caesar2020nuscenes}. 
There are six view records for samples of Nuscenes,~\ie, Front, Front Left, Front Right, Back Right, Back Left, and Back.
These views have some areas of overlap with one another.
In NuScenes, the collected data is annotated with a frequency of $2$Hz, and each annotated frame is referred to as a keyframe with annotations.

%
%
In our research, we propose an SQL-based approach for the automated generation of four types of instruction-follow data, namely: Perception, Prediction, Risk, and Planning with Reasoning. 
This methodology aligns with the sequential decision-making stages of human drivers, categorized as follows:
\textbf{1. Perception}: The initial stage of recognizing surrounding entities. \textbf{2. Prediction}: Forecasting the future actions of these entities. \textbf{3. Risk}: Identifying imminent dangers, such as vehicles executing overtaking manoeuvres. \textbf{4. Planning with Reasoning}: Developing a safe travel plan grounded in logical analysis.


%
The detailed process is shown in Fig.~\ref{fig:datageneration}.
Specifically, \textbf{1).} The \textbf{filter \& construction step} leverages the \textbf{(a)} original annotations to generate the scene information database (see Fig.~\ref{fig:datageneration}~{(b)}). 
\textbf{2).} The \textbf{sampling step} first samples three keyframes from the original dataset.
Then, as shown in Fig.~\ref{fig:datageneration}~{(c)}), we construct a series of pre-defined task SQLs.
Each task SQL consists of several subtasks, each of them consisting of a subtask function and an instruction prompt.
%
%
\textbf{3).} The \textbf{retrieval step} uses the instruction prompt and the task SQL to retrieve the corresponding response from the scene database.
\textbf{4).} The \textbf{saving step} saves all instruction-response pairs  (see Fig.~\ref{fig:datageneration}~{(d)}).
\textbf{5).} The \textbf{verifying step} employs either human analysis or LLM-based methods (e.g., GPT-4~\cite{OpenAI_2023}) to eliminate erroneous instruction-response pairs, thereby guaranteeing the quality of \dataset.
%
Our task SQL design is logically sequenced, and based on the inherent relational flow of autonomous driving tasks,~\ie, `Perception $\rightarrow$ Prediction, (Perception, Prediction) $\rightarrow$ Risk, (Risk, Prediction) $\rightarrow$ Planing with Reasoning', where $a \rightarrow b$ indicates the $b$ SQL is derived from the $a$ SQL (\textcolor{blue}{blue dashed arrows} in Fig.~\ref{fig:datageneration}~(c)).
We show a more detailed example for \textbf{Step 2} and \textbf{Step 3} in Fig.~\ref{fig:example}.
Specifically, three keyframes (\ie, from frame $i$ to frame $i+2$) with annotations are randomly sampled from the original dataset, and we only select one instance,~\ie, the pedestrian (\textcolor{red}{box}) for clarity (Fig.~\ref{fig:example}~(a)).
In this case, we choose distance, status, crossing, and planning subtask SQLs from the perception, prediction, risk, and planning with reasoning task SQLs, respectively (shown in Fig.~\ref{fig:example}~(b)).
The status subtask is based on the distance task, since the next status (\eg, speed, direction) for the instance is computed based on the distances of previous, current, and future frames. 
Each subtask SQL consists of two parts,~\ie, the subtask function and the instruction prompt.
For example, the distance subtask SQL has Find\_Distance($<$instance$>$) function and the instruction prompt is "What is the distance between $<$instance$>$ and the ego car? Please use the format as (x,y) where..", where \textcolor{red}{$<$instance$>$} is the input.
Finally, as shown in Fig.~\ref{fig:example}~(c), we use the instance information or frame information as the input for different subtask functions to retrieve the response from the scene database.
Compared with other data generation methods,~\eg, ChatGPT-based~\cite{xu2023drivegpt4} or human-based~\cite{dewangan2023talk2bev}, this structured design ensures the generation of instruction-response pairs is both reliable and scalable. 

We only describe the overview of the data generation in this section. Please refer to the supplementary material for more details about the scene information database, the task SQLs, and the retrieval process.

\if
The details of the four steps are as follows:
%
%
%

\vspace{0.5mm}
\noindent\textbf{Step 1 - Filter \& Construction.} 
Given a keyframe with annotations from NuScenes (Fig.~\ref{fig:datageneration}~(a)), we first compute the distances of all instances away from the ego car. Then we regard objects within $20$ meters from the ego car as important instances, while filtering others.
Based on the information of the ego car and important instances, we construct a scene database as shown in Fig.~\ref{fig:datageneration}~(b).
The scene database includes four tables to represent information about scenes, frames, ego cars, and other instances near the ego car.
%
%
Please refer to the supplementary material for more details about the scene database.
%

\vspace{0.2mm}
\noindent\textbf{Step 2 - Sampling.} After obtaining the scene database, we randomly sample three keyframes from one scene in the original dataset, which is then input to a specific task SQL sampled from a series of pre-defined task SQLs.
As shown in Fig.~\ref{fig:datageneration}~(c), we define four types of SQL sets (\ie, perception, prediction, risk, and planning with reasoning), which are later used for generating different types of instruction-response pairs.
Each type of task SQL set consists of several subtask SQLs.
For example, the perception SQL set contains six subtask SQLs,~\eg, Closest focuses on finding the objects closest to the ego car, and the other subtasks are similar.
%
%
For instance, the prediction SQLs are based on the perception ones, the risk SQLs are based on both prediction and perception SQLs, and the planning with reasoning SQL is inherited from prediction and risk ones.
%

\vspace{0.2mm}
\noindent\textbf{Step 3 - Retrieval.}
Given the sampled keyframes with annotations, we successively select subtask SQLs, each of them consisting of a subtask function and an instruction prompt.
Receiving the information of an instance or the ego car as input, the subtask function retrieves the responses from the scene information database (see Fig.~\ref{fig:datageneration}~(b)).
We show more detailed examples for step2 and 3 in Fig.~\ref{fig:example}.

\noindent\textbf{Step 4 - Saving.}
After obtaining the instruction prompts and their corresponding responses from \textbf{Step 3}, we save all different instruction-response pairs for different tasks.
Furthermore, we employ GPT-4~\cite{OpenAI_2023} to revise the responses for more fluent and reasonable.

We only describe the overview of the data generation in this section.
Please refer to the supplementary material for more details about the scene information database, the task SQLs, and the retrieval process.
\fi


\begin{figure*}[t]
    \centering
    \includegraphics[width=1.0\linewidth,height=0.25\textheight]{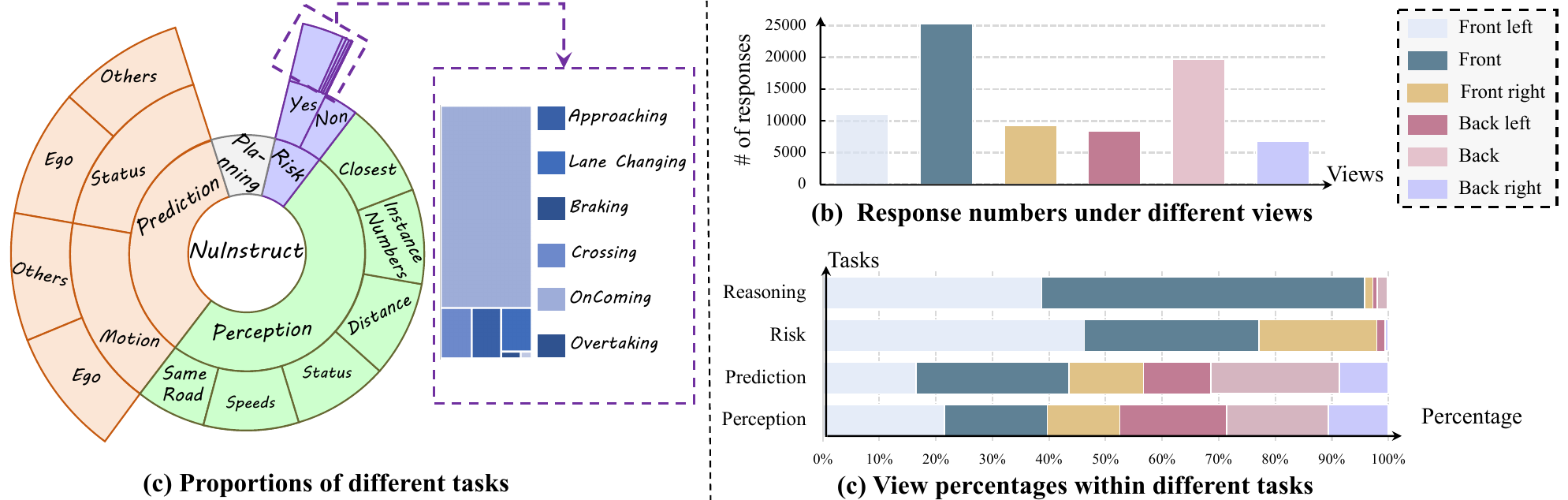}
    \vspace{-2.0em}
    \caption{\textbf{Statistics of NuInstruct}. \textbf{(a) Proportions of different tasks.} The size of the arc represents the proportions of each task, while the same color indicates tasks of the same category. Our task encompasses a diverse range of tasks including perception, prediction, risk, and planning. \textbf{(b) Response numbers under different views.}  The horizontal axis represents different views, and the vertical axis indicates the number of responses requiring information from the corresponding view. \textbf{(c) View percentages within different tasks.} The horizontal and vertical axes represent the proportion of different views and task classes, respectively.
    }
    \label{fig:statistics}
\end{figure*}

\subsection{Data Statistics}~\label{sec:datastatistics}
To construct our \dataset, we sampled a total of $11,850$ keyframes from $850$ videos within the NuScenes dataset~\cite{caesar2020nuscenes}. 
Subsequent filter yields $55,204$ unique instances, which collectively appear $295,828$ times across all keyframes. This culminates in an average of approximately $24.96$ instances per keyframe. 
%
By employing our SQL-based method (Section~\ref{sec:datagenerate}), we generated a total of $91,355$ instruction-response pairs, encompassing four primary tasks—namely, Perception, Prediction, Risk, and Planning with Reasoning. These tasks are further delineated into $17$ sub-tasks. The quantities of task categories are statistically presented in Fig.~\ref{fig:statistics}~(a).

Compared with other single-view benchmarks, our dataset covers multi-view information.
Hence, we also conduct a statistical analysis of the relations of different views and constructed instruction-response pairs.
Fig.~\ref{fig:statistics}~(b) shows the distribution of the numbers of the responses based on the views,~\ie, for a given view, we record how many responses are derived from information from the view.
Through such statistics, we find that to answer the instructions, the system needed to look at multiple views, instead of just a single one. 
%
In Fig.~\ref{fig:statistics}~(c), for each task, the proportions of responses obtained based on different views are calculated. We find two observations: \textbf{(i)} in the case of perception and prediction tasks, the distribution across views is relatively even, showing that our data generation method produces balanced multi-view information; \textbf{(ii)} When it comes to reasoning and risk tasks, the responses predominantly draw on information from the front, left-front, and right-front views. This is reasonable since drivers normally base their ahead or sides to decide the next actions, seldom needing to look behind.

\definecolor{mygray}{gray}{.9}

\begin{table}
    \centering
    \label{tab:rgr_with_and_without_artifacts}
    \begin{adjustbox}{max width=\linewidth}
    \begin{tabular}{l|c|c}
    \Xhline{1.5pt}
     \rowcolor{mygray}{\it \small \textbf{Task}}
    &{\small  \it \textbf{SubTask}}
    &{\small  \it \textbf{Metrics}}
    \\
    \midrule
    \multirow{2}{*}{Perception} &  Distance, Speeds, Instance Number & {MAE $\downarrow$}\\
    & Closest, Status, Same Road & Accuracy $\uparrow$\\
    \midrule
     \multirow{2}{*}{Prediction} &  Motion Ego, Motion Others & {MAE $\downarrow$}\\
     & Status Ego, Status Others & {Accuracy $\uparrow$}\\
    \midrule
   Risk & All &  MAP $\uparrow$\\
        \midrule
        Reasoning & All & BLEU~\cite{papineni2002bleu} $\uparrow$\\
    \bottomrule
    \end{tabular}
        \end{adjustbox}
            \vspace{-3.0mm}
         \caption{ \textbf{Evaluation metrics for different tasks.} $\downarrow$ represents the lower the scores, the better the results, while $\uparrow$ means the higher the scores, the better the results. `MAE' indicates the mean absolute error. `All' means the all subtasks.
    }
        \label{tab:metrics}
\end{table}

\subsection{Evaluation Protocols}~\label{sec:evaluationmetrics}

\vspace{-5mm}
\noindent \textbf{Evaluation Metrics.} 
Our \dataset~consists of diverse tasks, making it hard to evaluate the different tasks using one metric.
We summarize the evaluation metrics for different tasks in Table~\ref{tab:metrics}.
For tasks evaluated by MAE, we use the regular expression~\cite{friedl2006mastering} to obtain values.
More detailed computations for different metrics please refer to the supplementary material.

\noindent \textbf{Data Split.} Our \dataset~contains a total of $850$ videos from NuScenes~\cite{caesar2020nuscenes}. We split the all videos into training/validation/testing sets (7.5:1.5:1.5).
We train all models in the training set and select the model with the best performance in the validation set to report its results on the test set.


\section{Method}
%
In Section.~\ref{sec:preliminary}, we first give a preliminary for our framework,~\ie, input, output, task definition and notations.
Then, in Section~\ref{sec:baseline}, we provide Multi-view MLLM (MV-MLLM), a baseline that extends current multimodal large language models (MLLMs) for processing multi-view video inputs.
Finally, in Section~\ref{sec:bev-inmllm}, we propose BEV-InMLLM, which injects the bird's-eye-view (BEV) representations into current MLLMs for better panorama understanding for \dataset.

\subsection{Preliminaries}~\label{sec:preliminary}
Different from current MLLMs, the visual inputs for our model are
the multi-view videos $\{\mathbf{V}^i\}_{i=1}^{N_\text{view}}$, where $N_\text{view}$ is the total number of camera views, $\mathbf{V}^i = \{ \mathbf{v}^i_t  \}_{t=1}^{N_\text{frame}}$, $\mathbf{v}^i_t$ is the $t$-th frame in $\mathbf{V}^i$ and $N_\text{frame}$ is the total number of frames.
Instead of the predefined several tasks, we give a specific language instruction, we use a unified model to obtain its corresponding language response, as shown in Fig.~\ref{fig:teaser}.
For clarity in the following, we use $\mathbf{L}_\text{inst} \mathbb{R}^{N_\text{inst} \times D_\text{inst}}$ and $\mathbf{L}_\text{resp} \in \mathbb{R}^{N_\text{resp} \times D_\text{resp}}$ to denote the language instruction tokens and the response tokens respectively, which are generated by the language tokenizer~\cite{touvron2023llama}.
$N_\text{inst}$/$N_\text{resp}$ and $D_\text{inst}$/$D_\text{resp}$ are numbers of tokens and dimensions for the instruction/response.
%
\begin{figure}
    \centering
\includegraphics[width=1.0\linewidth,height=0.25\textheight]{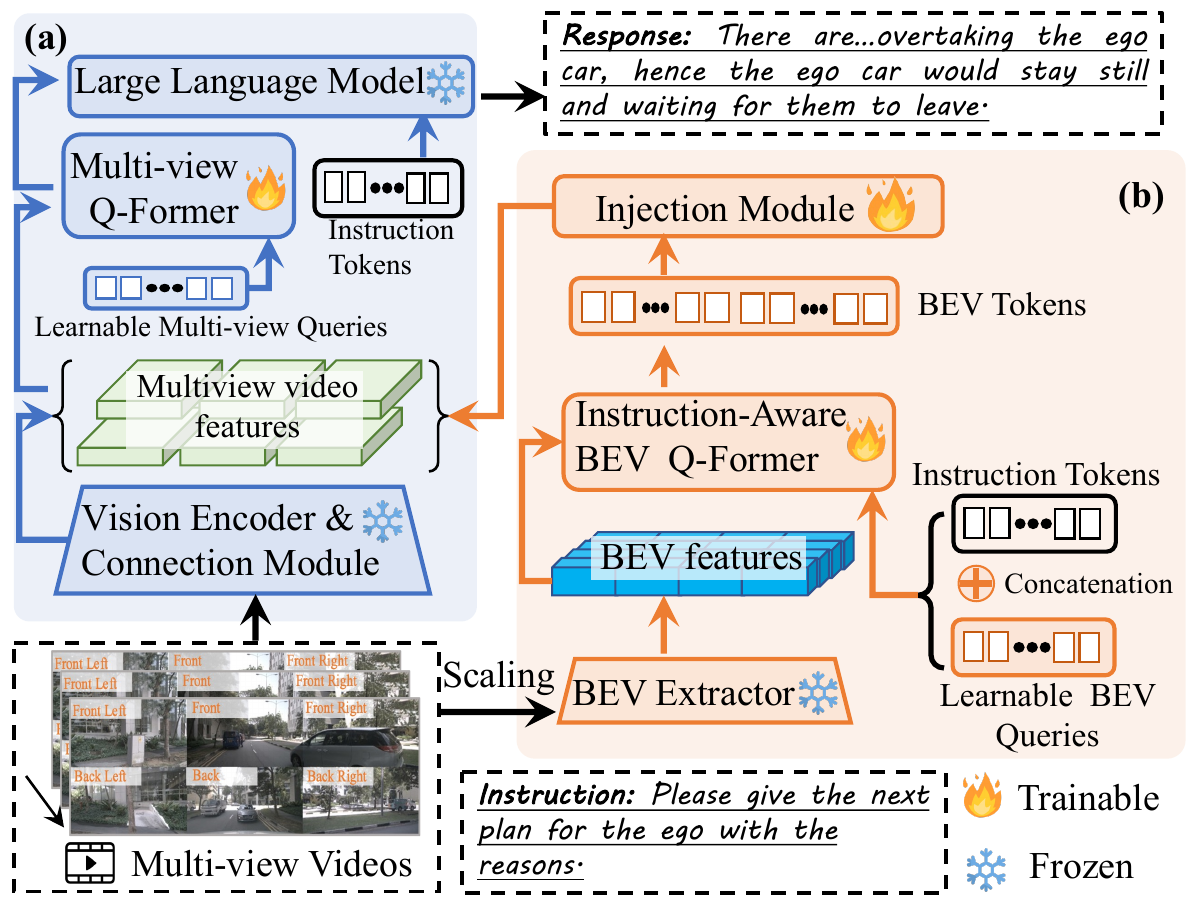}
\vspace{-2.0em}
    \caption{\textbf{The overall pipeline of our proposed BEV-InMLLM.} 
    \textbf{(a)} The base multimodal large language model (MLLM) tailored for processing the multi-view videos. 
    \textbf{(b)} The bird's-eye-view injection module (BEV-In) to inject BEV representations into base MLLM to boost \ad~understanding.
    %
    %
    }
    \label{fig:architecture}
\end{figure}
\subsection{Multi-View MLLM}~\label{sec:baseline}
Existing MLLMs~\cite{zhu2023minigpt,li2022blip,Dai2023instruct,li2023videochat} generally consist of three parts: a vision encoder $f_\text{vision}(\cdot)$ to receive the visual input; a connection module (\eg, Q-Former~\cite{li2023blip}) $f_\text{connect}(\cdot)$ to transfer the visual representations to the visual tokens aligned with the language; a large language model (LLM) $f_\text{LLM}(\cdot)$ to receive visual and language instruction tokens to generate the response.
Since they can only receive single-view input, we propose a baseline model named Multi-view MLLM (MV-MLLM) to enable the current MLLMs to process the multi-view videos, as shown in Fig.~\ref{fig:architecture}~(a).
Specifically, for a video from a specific view,~\ie, $\mathbf{V}^i$, we feed it into the vision encoder followed by the connect module to obtain the visual tokens, which can be formulated as:
\begin{equation}
    \mathbf{F}^i_\text{vis} = f_\text{connect}(f_\text{vision}(\mathbf{V}^i))  \in \mathbb{R}^{N_\text{vis} \times D_\text{vis}},
\end{equation}
where $N_\text{vis}$ and $D_\textbf{v}$ are the numbers of the visual tokens and the dimensions respectively.
Then, we introduce a multi-view Q-Former (similar to BLIP-2~\cite{li2022blip}) to capture the MV visual semantics $\mathbf{F}_\text{mv}$ from $\{ \mathbf{F}^i_\text{vis} \}_{i=1}^{N_\text{view}}$.
We concatenate $\{ \mathbf{F}^i_\text{vis} \}_{i=1}^{N_\text{view}}$ along the view dimension to obtain $\overline{\mathbf{F}}_\text{mv} \in \mathbb{R}^{(N_\text{view}*N_\text{vis}) \times D_\text{vis}}$.
The input to the multi-view Q-Former contains a set of $K_\text{mv}$ learnable multi-view queries $\mathbf{Q}_\text{mv} \in \mathbb{R}^{K_\text{mv} \times D_\text{vis}}$, which interact with $\overline{\mathbf{F}}_\text{mv}$ through the cross attention, formulated as follows:
\begin{equation}
    \mathbf{F}_\text{mv} = \text{CrossAttn}(\mathbf{Q}_\text{mv}, \overline{\mathbf{F}}_\text{mv})  \in \mathbb{R}^{K_\text{mv} \times D_\text{vis}}.
    \label{E:mvqformer}
\end{equation}

%
The output $\mathbf{F}_\text{mv}$ then goes through a linear projection (omitted in Fig.~\ref{fig:architecture}), and is fed to the LLM.
Note that in our MV-MLLM, only the MV Q-Former is trainable and other parameters are frozen to fully retain the knowledge of the pre-trained models.


\subsection{BEV-Injected MLLM}~\label{sec:bev-inmllm}
%
The BEV approach has become pivotal in autonomous driving for its precise depiction of object positioning, essential for tasks like perception and planning~\cite{ma2022vision,li2022bevformer}. Integrating BEV into our MV-MLLM offers enhanced visual-spatial analysis for \ad.
%
While BEV can be constructed from the multi-view features $\{ \mathbf{F}^i_\text{vis} \}_{i=1}^{N_\text{view}}$ by physical transformations such as LSS~\cite{philion2020lift}, similar to NuScenes-QA~\cite{qian2023nuscenes}, the plain of ViTs in current MLLMs limits their perception capabilities~\cite{pan2022parameter}.
Replacing the VIT with BEV-specific backbones,~\eg, ResNet~\cite{He_Zhang_Ren_Sun_2016} or Swin Transformer~\cite{liu2021swin}, diminishes visual-language alignment~\cite{ding2023hilm}.
Furthermore, the limited input resolutions generate small feature maps, which are hard to scale up to the high-resolution BEV representation.
%
%

%
%
%

To address the above problems, we propose the BEV-injected MLLM (BEV-InMLLM), which uses a BEV injection module (BEV-In) to obtain the BEV information aligned with LLMs in a data-efficient and resource-light way.
As shown in Fig.~\ref{fig:architecture}~(b), we obtain the high-quality BEV features $\mathbf{F}_\text{bev} \in \mathbb{R}^{W \times H \times D_\text{bev}}$ from the pre-trained BEV extractor~\cite{philion2020lift,huang2021bevdet}, where $W$, $H$ and $D_\text{bev}$ denote the width, height and dimensions. The following are two key components of BEV-In,~\ie, an instruction-aware BEV Q-Former and an injection module.

\noindent\textbf{Instruction-aware BEV Q-Former.}  We introduce the instruction-aware BEV Q-Former to ignore the redundant and irrelevant to the given instructions from $\mathbf{F}_\text{bev}$.
The input queries for the instruction-aware BEV Q-Former blend two parts: the instruction tokens $\mathbf{L}_\text{inst}$ for instruction-related aspects, and the learnable BEV queries for extracting the useful information pertinent to the instruction from $\mathbf{F}_\text{bev}$.
The process of the instruction-aware BEV Q-Former is defined as:
\begin{equation}
    \mathbf{F}_\text{instbev} = \text{CrossAttn}(\mathbf{Q}_\text{bev} \bigoplus \mathbf{L}_\text{inst}, \mathbf{F}_\text{bev})
    \label{E:bevquery}
\end{equation}
where $\bigoplus$ indicates the concatenation, $\mathbf{Q}_\text{bev} \in \mathbb{R}^{K_\text{bev} \times D_\text{vis}}$ and $K_\text{bev}$ indicates and the BEV queries and their the numbers, $\mathbf{F}_\text{instbev} \in  \mathbb{R}^{(K_\text{bev} + N_\text{inst}) \times D_\text{vis}}$ are the instruction-aware BEV tokens.

\noindent\textbf{Injection module.} 
Our injection module fuses multi-view features $\overline{\mathbf{F}}\text{mv}$ with instruction-aware BEV tokens $\mathbf{F}\text{instbev}$ through cross-attention:
\begin{equation}
    \overline{\mathbf{F}}_\text{mv} = \overline{\mathbf{F}}_\text{mv} + \text{CrossAttn}(\overline{\mathbf{F}}_\text{mv}, \mathbf{F}_\text{instbev}),
    \label{E:inject}
\end{equation}
%
where the enhanced $\overline{\mathbf{F}}_\text{mv}$ contains both (i) temporal multi-view cues for scene comprehension and (ii) spatial-aware BEV information for precise perception and planning tasks. We keep our BEV-In module efficient by making only two components trainable: the BEV Q-Former and the injection module, while the BEV feature extractor remains frozen to maintain feature quality.

\begin{table*}
    \centering
    \label{tab:sargd_ablation_study}
    \begin{adjustbox}{max width=\textwidth}
    \begin{tabular}{l|c c c c c c | c c | c c c c c c | c}
    \Xhline{1.5pt}
    \rowcolor{mygray}
    &\multicolumn{6}{c|}{\it \textbf{Perception}}
    &\multicolumn{2}{c|}{\it \textbf{Prediction}}
    &\multicolumn{6}{c|}{\it \textbf{Risk}}
    &\\
    \rowcolor{mygray}
    \multirow{-2}{*}{\textbf{Method}} &  Dis$\downarrow$ &  Sped $\downarrow$ & \# Ins $\downarrow$  & Clos $\uparrow$   & Sta $\uparrow$ & SameR $\uparrow$
    & Mot $\downarrow$ & Sta $\uparrow$
    & App $\uparrow$ & LaneC $\uparrow$ & Onco $\uparrow$&  Cro $\uparrow$  & Over $\uparrow$  & Brak $\uparrow$  &  \multirow{-2}{*}{\thead{\it \textbf{Planning} \\ \textbf{with  Reasoning} } $\uparrow$ }
     
    \\
    \hline \hline
    BLIP-2$^*$~\cite{li2022blip} &29.3 & 5.6 & 4.4 & 18.5 & 15.9 & 23.8 & 8.7 & 38.7 & 6.1 & 10.6 & 15.4 & 16.7 & 3.7 & 21.5 & 24.9\\
                    MV-MLLM  & 26.8 & 5.3 & 3.9 & 28.2 & 17.7 & 31.5 & 6.8 & 43.6 & 15.2 & 19.3  & 18.4 & 22.7& 9.6 & 22.7 & 30.1 \\
                 {BEV-InMLLM}  & \textbf{23.3} & \textbf{3.6} & \textbf{3.2} & \textbf{33.6} & \textbf{18.9} & \textbf{31.6} & \textbf{3.8} & \textbf{45.2} &  \textbf{16.8} & \textbf{21.0}  & \textbf{19.7} & \textbf{23.9} &\textbf{10.5} & \textbf{27.5} & \textbf{33.3}\\
    \midrule
MiniGPT-4$^*$~\cite{zhu2023minigpt} & 30.2 & 6.2 & 6.3 & 20.2	&17.3 & 24.2 &	8.7 & 39.6 & 7.8 & 12.5
                              & 16.9 & 18.7& 4.8 & 21.8 &26.3\\
                    MV-MLLM  & 28.6 & 4.7 & 4.1 & 27.5 & 18.5 & 30.7 & 7.2 & 44.2 & 15.5 & 18.9              & 19.1 & 23.3& 8.2 & 23.1 & 32.3 \\
                BEV-InMLLM& \textbf{23.6} & \textbf{3.1} & \textbf{3.8} & \textbf{32.9} & \textbf{19.2} & \textbf{31.5} & \textbf{4.2} & \textbf{46.5} &  \textbf{17.3} & \textbf{20.5}                & \textbf{21.5} & \textbf{24.5}& \textbf{9.4} & \textbf{26.8} & \textbf{35.6}\\
    \midrule
    Video-LLama~\cite{zhang2023video} & 29.9 & 6.5 & 5.4 & 22.3 & 16.7 & 20.9 & 9.3 & 39.3 & 6.2 & 10.9 & 16.2 & 18.4 & 4.1 & 21.3 & 25.3\\              
      MV-MLLM & 28.9 & 6.2 & 2.4 & 27.9 & 19.6 & 30.9 & 9.3 & 44.3 & 16.5 & 18.7 & 19.9 & 23.0 & 6.5 & 26.6 & 31.4\\
    {BEV-InMLLM} & \textbf{24.5} & \textbf{3.5} & \textbf{4.2} & \textbf{31.6} & \textbf{19.0} & \textbf{34.6} & \textbf{4.1} & \textbf{44.7} & \textbf{17.7} & \textbf{22.5} & \textbf{21.4} & \textbf{26.1} & \textbf{8.7} & \textbf{27.9} & \textbf{35.2}\\
    \bottomrule
    \end{tabular}
    \end{adjustbox}
     \vspace{-3.0mm}
     \caption{\textbf{Performance comparison with state-of-the-arts on \dataset.} Optimal scores are highlighted in bold. Note that all models are fine-tuned on the training set of \dataset~in the same setting. `$^*$' means we use the spatiotemporal adapter to enable the image-based MLLM to receive the video input. For clarity, we employ abbreviations to denote the names of subtasks instead of their full designations,~\ie, Dis = Distance, Sped = Speeds, \# Ins = Instance number, Clos = Closest, Sta = Status, SameR = In the same road, Mot = Motion, App = Approach, LaneC = Lane changing, Onco = On coming, Cro = Crossing, Over = Overtaking, Brak = Braking. Best results are reported in \textbf{Bold}.
    }
    \label{tab:sota}
\vspace{-1.0em}
\end{table*}

\vspace{-2mm}
\section{Experiments}
%

\vspace{-2mm}
\noindent\textbf{Implementation and training details.}
We experiment on three base MLLMs,~\ie 
We evaluated our MV-MLLM and BEV-InMLLM on three base MLLMs: BLIP2~\cite{li2023blip}, Video-LLama~\cite{zhang2023video}, and MiniGPT-4~\cite{zhu2023minigpt}. 
To adapt BLIP2 and MiniGPT-4, which are image-centric, for video input, we used the spatiotemporal adapter (ST-Adapter)\cite{pan2022parameter}, following\cite{ding2023hilm}, while preserving their pre-trained parameters.
%
%
We initialized all MLLMs with their official pre-trained weights~\footnote{\url{https://github.com/salesforce/LAVIS/};\url{https://github.com/Vision-CAIR/MiniGPT-4}; \url{https://github.com/DAMO-NLP-SG/Video-LLaMA}}, freezing these during training and only training the parameters of ST-Adapters and our additional modules (MV Q-Former, BEV Q-Former, and injection module).
We choose LSS~\cite{philion2020lift} and BEVFormer~\cite{li2022bevformer} as our BEV extractors, and $W$ and $H$
are both set to $200$.
$N_\text{view}$, $K_\text{mv}$, $K_\text{bev}$ are set to $6$, $32$ and $32$ respectively.
The dimensions,~\ie, $D_\text{vis}$ and $D_\text{resp}$ are both set to $1408$, the same as the dimension of the same as EVA\_CLIP hidden feature dim used by BLIP-2.
The input is resized and cropped to the spatial size of $224 \times 224$, and each video is uniformly sampled $3$ frames.
%
%
%
We use AdamW~\cite{loshchilov2017decoupled} as the optimizer and cosine annealing scheduler~\cite{loshchilov2016sgdr} as the learning rate scheduler with an initial learning rate of $1e$-$4$.
We train all models in a total of $20$ epochs.
%

%

\subsection{State-of-the-art Comparison}~\label{sec:sota}
We select three advanced MLLMs,~\ie, BLIP-2~\cite{li2023blip}, MiniGPT-4~\cite{zhu2023minigpt} and Video-LLama~\cite{zhang2023video} as our base models.
For each MLLM, we apply our proposed modules to obtain MV-MLLM and BEV-InMLLM.
All models are finetuned in the same setting.
We report our results on \dataset~test set in Table~\ref{tab:sota}.
%
%
To conserve space, we aggregate the reporting of two subtasks, `motion ego' and `motion others'. A similar approach is adopted for `status ego' and `status others'.

From Table~\ref{tab:sota}, we observe that equipped with our proposed modules, there is a significant increase in the evaluation metrics on all tasks, demonstrating its effectiveness. 
%
More specifically, the integration of temporal and multi-view information (MV-MLLM) substantially improves risk and planning tasks by $~5\%$ and $~6\%$, respectively.
%
Furthermore, injecting BEV into MV-MLLM,~\ie, BEV-InMLLM, benefits tasks sensitive to distance and position,~\eg, perception and prediction.

\definecolor{mygray}{gray}{.9}

\begin{table}
    \centering
    \label{tab:rgr_with_and_without_artifacts}
    \begin{adjustbox}{max width=\linewidth}
    \begin{tabular}{l|c|c|c|c}
    \Xhline{1.5pt}
    \rowcolor{mygray} {\it \small \textbf{Task}}
    &{\small  \it \textbf{Temporal}}
    &{\small  \it \textbf{Multi-view}}
    &{\small  \it \textbf{Spatial}}
    &{\small \it \textbf{Holistic}}
    \\
    \midrule
    {\it \small {SubTask}}
    & Sped, Mot, Sta
    & \# Ins, SameR 
    & Dis, Mot, Sta, Clos
    & Risk, Planning\\
    \bottomrule
    \end{tabular}
        \end{adjustbox}
        \vspace{-3.0mm}
         \caption{\textbf{Reclassified tasks.} To better analyze the impact of each module on autonomous driving tasks, we reclassify the sub-tasks into four main tasks based on their dependency on different types of information. 
    `\#' indicates the numbers. `Holistic' means those tasks that require all information,~\ie, temporal, multi-view, and spatial.
    }
        \label{tab:categorytask}
\end{table}

\begin{table}
    \centering
    \begin{adjustbox}{max width=\linewidth}
    \begin{tabular}{l|cc|cc|cc|c}
    \Xhline{1.5pt}
    \rowcolor{mygray}
    &\multicolumn{2}{c|}{\small  \it \textbf{Temporal}}
    &\multicolumn{2}{c|}{\small  \it \textbf{Multi-view}}
    &\multicolumn{2}{c|}{\small  \it \textbf{Spatial}}
    &
    \\
    \rowcolor{mygray}
     \multirow{-2}{*}{\small \textbf{Model}}
    &\textbf{$\downarrow$}
    &\textbf{$\uparrow$}
     &\textbf{$\downarrow$}
    &\textbf{$\uparrow$}
   &\textbf{$\downarrow$}
    &\textbf{$\uparrow$}
    & \multirow{-2}{*}{\small \it \textbf{Whole} $\uparrow$}\\
  \midrule
   \textbf{(a)}  Full  & 3.7 & 32.8 & 3.8 & 31.5 &13.9&32.9 & 22.2  \\
    \midrule
\textbf{(b)} w/o Video & 7.4 & 28.5 & 4.2 & 30.4 & 14.2 & 30.8 & 21.0 \\
$\triangle$ & \underline{\textbf{\textcolor{mygreen}{-3.7}}} & \underline{\textbf{\textcolor{mygreen}{-4.3}}} & \transparent{0.6}\textbf{\textcolor{mygreen}{-0.4}} & \transparent{0.6}\textbf{\textcolor{mygreen}{-1.1}} & \transparent{0.6}\textbf{\textcolor{mygreen}{-0.3}} & \transparent{0.6}\textbf{\textcolor{mygreen}{-1.1}} & \transparent{0.6}\textbf{\textcolor{mygreen}{-1.2}} \\
   \midrule
   \textbf{(c)} w/o MV & 5.2 & 31.2 & 6.0 & 28.3 & 14.4 & 32.5 & 21.6 \\
$\triangle$ & \transparent{0.6}\textbf{\textcolor{mygreen}{-1.5}} & \transparent{0.6}\textbf{\textcolor{mygreen}{-1.6}} & \underline{\textbf{\textcolor{mygreen}{-2.2}}} & \underline{\textbf{\textcolor{mygreen}{-3.2}}} & \transparent{0.6}\textbf{\textcolor{mygreen}{-0.5}} & \transparent{0.6}\textbf{\textcolor{mygreen}{-0.4}} & \transparent{0.6}\textbf{\textcolor{mygreen}{-0.6}} \\
   \midrule
    \textbf{(d)} w/o BEV & 6.0 & 31.4 & 4.1 & 30.7 & 18.0 & 30.0 & 20.1 \\
$\triangle$ & \underline{\textbf{\textcolor{mygreen}{-2.3}}} & \transparent{0.6}\textbf{\textcolor{mygreen}{-1.4}} & \transparent{0.6}\textbf{\textcolor{mygreen}{-0.3}} & \transparent{0.6}\textbf{\textcolor{mygreen}{-0.8}} & \underline{\textbf{\textcolor{mygreen}{-4.1}}} & \underline{\textbf{\textcolor{mygreen}{-2.9}}} & \underline{\textbf{\textcolor{mygreen}{-2.1}}} \\
   \midrule
   \textbf{(e)} Base & 10.3 & 25.8 & 6.7 & 22.7 & 20.8 & 27.6 & 12.4 \\
$\triangle$ & \underline{\textbf{\textcolor{mygreen}{-6.6}}} & \underline{\textbf{\textcolor{mygreen}{-7.0}}} & \underline{\textbf{\textcolor{mygreen}{-2.9}}} & \underline{\textbf{\textcolor{mygreen}{-8.8}}} & \underline{\textbf{\textcolor{mygreen}{-6.9}}} & \underline{\textbf{\textcolor{mygreen}{-5.3}}} & \underline{\textbf{\textcolor{mygreen}{-9.8}}} \\
  \bottomrule
    \end{tabular}
    \end{adjustbox}
\label{tab:effectofmodule}
 \vspace{-3.0mm}
 \caption{\textbf{The ablation study of different proposed modules.} `w/o' indicates the without the specific module.
    `Video', `MV' and `BEV' indicate video input, MV Q-Former (Section~\ref{sec:baseline}) and BEV injection module (Section~\ref{sec:bev-inmllm}) respectively.
    The results of performance degradation exceeding $2$ are reported in \underline{\textbf{\textcolor{mygreen}{green}}}.
    $\triangle$ is the difference between a specific model with full model,~\ie, line (a).
    %
    }
    \label{tab:effectofmodule}
\end{table}

\begin{table*}[t]
\vspace{-.2em}
\centering
\subfloat[
\textbf{BEV extractor}. We select three advanced BEV extractors. Powerful extractors are more effective.
\\{\scriptsize $^\dagger$: using additional lidar-modal data.}
\label{tab:bevextractor}
]{
\begin{minipage}{0.24\linewidth}{\begin{center}
\tablestyle{6pt}{1.05}
\begin{tabular}{y{45}x{25}x{24}}
extractor & {\it \textbf{Spatial}}$\downarrow$ & {\it \textbf{Whole}}$\uparrow$ \\
\shline
LSS~\cite{philion2020lift} & \baseline{13.9} & \baseline{21.0} \\
BEVDet~\cite{huang2021bevdet} & {13.6} & {21.3} \\
BEVFusion~\cite{liu2023bevfusion}$^\dagger$ &\textbf{13.2} & \textbf{21.5} \\
\end{tabular}
\end{center}}\end{minipage}
}
\hspace{1em}
\subfloat[
\textbf{Instruction tokens $\mathbf{L}_\text{inst}$}. `w/o' and `w/' indicate without and with. $\mathbf{L}_\text{inst}$ (Eq.~\ref{E:bevquery}) can extract the instruction-aware BEV. 
\label{tab:instructiontoken}
]
{
\begin{minipage}{0.2\linewidth}{\begin{center}
\tablestyle{1pt}{1.05}
\begin{tabular}{y{15}z{30}z{30}}
$\mathbf{L}_\text{inst}$ & {\it \textbf{Spatial}}$\downarrow$ & {\it \textbf{Whole}}$\uparrow$ \\
\shline
w/o & 15.3 & 20.1  \\
w/ & \baseline{\textbf{13.9}} & \baseline{\textbf{21.5}}\\
\multicolumn{3}{c}{~}\\
\end{tabular}
\end{center}}\end{minipage}
}
\centering
\hspace{1em}
\subfloat[
\textbf{BEV query number $K_\text{bev}$}. More numbers for BEV queries $\mathbf{Q}_\text{bev}$ (Eq.~\ref{E:bevquery}) benefit the model.
\label{tab:querynumber}
]{
\centering
\begin{minipage}{0.18\linewidth}{\begin{center}
\tablestyle{4pt}{1.05}
\begin{tabular}{y{15}x{30}x{22}}
$K_\text{bev}$ &{\it \textbf{Spatial}$\downarrow$ }& {\it \textbf{Whole}}$\uparrow$ \\
\shline
16 & 14.5 & 21.0 \\
32 & \baseline{\textbf{13.9}} & \baseline{{21.5}} \\
64 & \textbf{13.9} & \textbf{21.6} \\
\end{tabular}
\end{center}}\end{minipage}
}
\hspace{1em}
\subfloat[
\textbf{Injection feature}. BEV features $\mathbf{F}_\text{insbev}$ can be injected into $\mathbf{F}_\text{mv}$ or $\overline{\mathbf{F}}_\text{mv}$ (Eq.~\ref{E:inject}). The latter is better due to more information.
\label{tab:injectfeat}
]{
\begin{minipage}{0.2\linewidth}{\begin{center}
\tablestyle{1pt}{1.05}
\begin{tabular}{y{25}x{30}x{30}}
feature & {\it \textbf{Spatial}} $\downarrow$& {\it \textbf{Whole}}$\uparrow$  \\
\shline
$\mathbf{F}_\text{mv}$  & 15.1 & 20.8  \\
$\overline{\mathbf{F}}_\text{mv}$& \baseline{\textbf{13.9}} & \baseline{\textbf{21.5}} \\
\multicolumn{3}{c}{~}\\
\end{tabular}
\end{center}}\end{minipage}
}
\vspace{-.9em}
\caption{\textbf{BEV Injection module ablation experiments} with MiniGPT-4 on \dataset. Best results are reported in \textbf{Bold}. Default settings are marked in \colorbox{baselinecolor}{gray}.}
\label{tab:ablations} \vspace{-.5em}
\end{table*}

\subsection{Ablation Study}~\label{sec:ablation}
In this section, we conduct experiments to evaluate the effect of different proposed modules and different input information.
Here, we use MiniGPT-4~\cite{zhu2023minigpt} as our baseline model.
%
%
To better analyze the impact of each module on autonomous driving tasks, we reclassify the sub-tasks into four main tasks based on their dependency on different types of information. These are categorized as temporal-related (temporal), multi-view-related (multi-view), spatial-related (Spatial), and holistic-related tasks, as shown in Table~\ref{tab:categorytask}.
`Temporal' indicates subtasks related to temporal cues,~\eg, the vehicle's status is determined based on its positions at various times, and so do others.
Note that some subtasks may be classified into different tasks,~\eg, the status task is in both temporal and spatial tasks.
We will report the results of different models under the reclassified tasks in the following.

\vspace{-3.5mm}
\subsubsection{Effect of different proposed modules}
\vspace{-2.5mm}
In our study, we explore different modules to capture different information for autonomous driving tasks: ST-Adapter accepts videos for temporal, MV-Q Former for multi-view, and BEV Injection module for location, distance, and spatial information in BEV features. We use BEV-InMLLM as a full model including comprehensive information types, then sequentially remove each module to derive the following distinct models:
\textbf{(a)} The full model,~\ie, BEV-InMLLM introduced in Section~\ref{sec:bev-inmllm}. \textbf{(b)} BEV-InMLLM without temporal cues,~\ie, the input is image. \textbf{(c)} BEV-InMLLM without multi-view information,~\ie, only single-view input. \textbf{(d)} BEV-InMLLM without BEV information,~\ie, without BEV injection module. \textbf{(e)} The baseline model,~\ie, MiniGPT-4~\cite{zhu2023minigpt}.

We report the results of different models in Table~\ref{tab:effectofmodule}.
From the table, we observe the following findings:
\textbf{(i)} Compared with (a) and (b), without temporal information, the performance of tasks highly dependent on temporal cues would degrade clearly, proving the importance of video input. We can also observe a similar phenomenon when comparing the results with (a) and (c).
\textbf{(ii)} Information contained in BEV is very important for most of \ad~tasks, since it clearly presents the surroundings of the ego vehicle, thus aiding the model in making informed decisions.

\vspace{-5mm}
\subsubsection{Analysis of BEV Injection Module}
\vspace{-2.5mm}
We ablate our BEV injection module (BEV-In) (Fig.~\ref{fig:architecture}~(b)) using the default settings in Table~\ref{tab:ablations} (see caption). Several intriguing properties are observed.

\paragraph{BEV extractor.}
We compare the performance of different BEV extractors in Table~\ref{tab:bevextractor}. Our results show that more strong extractor,~\eg, BEVDet~\cite{huang2021bevdet}, outperforms the weak one LSS~\cite{philion2020lift}.
Furthermore, BEVFusion~\cite{liu2023bevfusion} uses RGB and lidar modality for best performance.
Here, we use RGB images for efficiency.

\paragraph{Intruction tokens $\mathbf{L}_\text{inst}$ in BEV-In.}
Table~\ref{tab:instructiontoken} studies the effect of  $\mathbf{L}_\text{inst}$ in Eq.~\ref{E:bevquery}.
`w/o' indicate only using $\mathbf{Q}_\text{bev}$ to interact with $\mathbf{F}_\text{bev}$.
Results show that using instruction tokens can capture more related BEV features, thus improving the performance by $1.4$ for both spatial tasks and holistic tasks.

\paragraph{BEV query number $K_\text{bev}$}. In Table~\ref{tab:querynumber}, we study the influence of BEV query numbers,~\ie, $K_\text{bev}$ in $\mathbf{Q}_\text{bev}$ (Eq.~\ref{E:bevquery}).
As the number increases, the performance would be improved,~\eg, $0.6$ on the spatial performance with $K_\text{bev}$ arise from $16$ to $32$.
Considering setting $K_\text{bev}$ to $64$ only brings a small improvement, we use $32$ as the default settings for computation efficiency.

\paragraph{Injection feature}. The key design of our BEV-InMLLM is to inject instruction-aware BEV features (~\ie, $\mathbf{F}_\text{insbev}$ in Eq.~\ref{E:inject}) to the MV-MLLM.
In Table~\ref{tab:injectfeat}, we compare the performance of different features to inject with the $\mathbf{F}_\text{insbev}$.
Specifically, the multi-view visual semantics $\mathbf{F}_\text{mv}$ (Section~\ref{sec:baseline}) and the output of multi-view Q-Former $\overline{\mathbf{F}}_\text{mv}$ (Eq.~\ref{E:mvqformer}).
%
We find injecting into $\overline{\mathbf{F}}_\text{mv}$ achieves better, $0.7\%$ improvement over $\mathbf{F}_\text{mv}$ on the holistic tasks.
The reason is that $\mathbf{F}_\text{mv}$ is the filtered visual tokens, losing much spatial information.
%

\section{Conclusion}
In this study, we investigate language-based driving for \ad tasks. We introduce \dataset, featuring 91K multi-view video-instruction-response pairs across 17 subtasks, created via a novel SQL-based method. Our proposed BEV-InMLMM integrates instruction-aware BEV features into MLLMs, enhancing temporal, multi-view, and spatial detail processing.
BEV-InMLMM, as a plug-and-play enhancement, boosts MLLM performance on \ad tasks. Our empirical results on \dataset~confirm our method's efficacy.

\noindent\textbf{Limitations.} The current dataset lacks traffic light information and tasks related to 3D object detection, which we plan to address in future work.


{
    \small
    \bibliographystyle{ieeenat_fullname}
    \bibliography{main}
}
\clearpage
\setcounter{page}{1}
\maketitlesupplementary

\section{More details about \dataset}
\begin{table*}[t!]
     \begin{subtable}{0.58\columnwidth}
		\vspace{1.5mm}
	\resizebox{1.0\columnwidth}{!}{
		\begin{tabular}{   c  | c c }
    \Xhline{1.5pt}
     \rowcolor{mygray} Field &{\it \small \textbf{Scene ID}}
    &{\small  \it \textbf{Frame ID List}}\\
  \Xhline{1.5pt}
			 Type & string
    & string list
    \\
    \bottomrule
	\end{tabular}}
	\label{tab:sceneinfo}
 \vspace{-1.5mm}
  \caption{{\bf Scene Information Table $T_\text{scene}$}. }
    \end{subtable}
    \begin{subtable}[t]{0.77\columnwidth}
    \end{subtable}
     \begin{subtable}[t]{0.66\columnwidth}
    \end{subtable}
    \vspace{1.5mm}
     \begin{subtable}{1.3\columnwidth}
	\resizebox{0.9\columnwidth}{!}{
		\begin{tabular}{   c  | c c  c  }
   \Xhline{1.5pt}
			 \rowcolor{mygray} Field &{\it \small \textbf{Frame ID}}
    &{\small  \it \textbf{Ego Information ID List}} & {\small  \it \textbf{Instance Information ID List}} \\
			\Xhline{1.5pt}
			  Type & string token
    & string list & string  list\\
    \bottomrule
	\end{tabular}
	}
	\label{tab:frameinfo}
 \vspace{-1.mm}
 \caption{{\bf Frame Information Table $T_\text{frame}$}.}
    \end{subtable}
    \vspace{1.5mm}
     \begin{subtable}{2.0\columnwidth}
     \centering 
	\resizebox{0.8\columnwidth}{!}{
		\begin{tabular}{   l | c c  c c c c  }
   \Xhline{1.5pt}
			  \rowcolor{mygray} Field &{\it \small \textbf{Information ID}}
    &{\small  \it \textbf{Pose}} & {\small  \it \textbf{Rotation}} & {\small  \it \textbf{Velocity}} & {\small  \it \textbf{Road Information}} & {\small  \it \textbf{Camera Information}}
    \\
    \midrule
    Type & string & float list & float list & float & dictionary & dictionary\\
    \bottomrule
	\end{tabular}
	}
	\label{tab:egoinfo}
 \vspace{-1.mm}
 \caption{{\bf Ego Information Table $T_\text{ego}$}.}
    \end{subtable}
     \begin{subtable}{2.0\columnwidth}
     \centering
	\resizebox{1.0\columnwidth}{!}{
		\begin{tabular}{   l | c c  c c c c c c c c c}
   \Xhline{1.5pt}
			  \rowcolor{mygray} Field &{\it \small \textbf{Information ID}}&{\it \small \textbf{Instance ID}} & {\small  \it \textbf{Category}} & {\small  \it \textbf{Attribute}}
    &{\small  \it \textbf{Global-T}} & {\small  \it \textbf{Global-R}} &{\small  \it \textbf{Local-T}} & {\small  \it \textbf{Local-R}} & {\small  \it \textbf{Velocity}} & {\small  \it \textbf{Road Information}} & {\small  \it \textbf{Camera Pos}} 
    \\
    \midrule
    Type & string
    & string & string& string &float list & float list & float list & float list & float & dictionary & dictionary \\
    \bottomrule
	\end{tabular}
	}
	\label{tab:instanceinfo}
  \vspace{-1.mm}
    \caption{{\bf Instance Information Table $T_\text{ins}$}. Global-T = Global Translation, Global-R = Global Rotation, Local-T = Local Translation, Local-R = Local Rotation}.
    \end{subtable}
     \vspace{-3.mm}
     \caption{\textbf{Detaild Field and Type for different tables in} Fig.~\ref{fig:databasesql}~(a)}.
    \label{tab:database}
\end{table*}

In this section, we give more information about our \dataset.
In Section~\ref{sec:database}, we show the detailed information of the scene database.
Then, we give the definition of Algorithms~\ref{alg:distance}-\ref{alg:planning} for all task SQLs in Section~\ref{sec:sql}.
Finally, the computation for different metrics is presented in Section~\ref{sec:evaluation}.

\begin{figure}
    \centering
    \includegraphics[width=\linewidth,height=0.33\textheight]{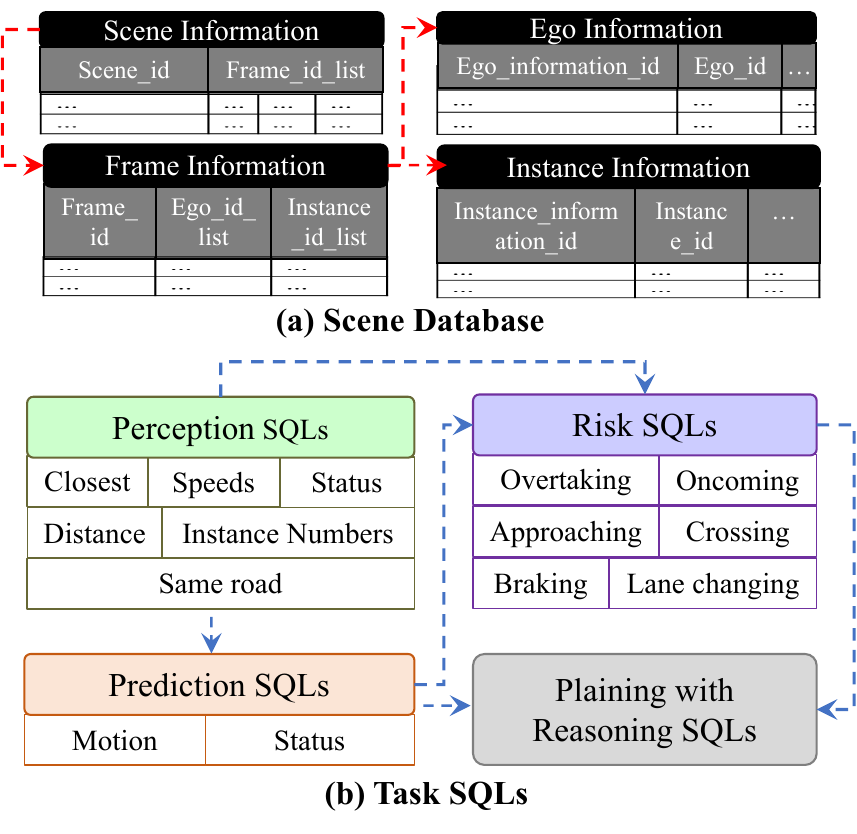}
    \caption{The illustration of \textbf{(a) Scene Database} and \textbf{(b) Task SQLs.} The \textcolor{red}{red dashed line} indicates the mapping relation of different tables. The \textcolor{blue}{red blue line} indicates the derivation.
    }
    \label{fig:databasesql}
\end{figure}



\subsection{Scene Database}~\label{sec:database}

%
Based on the information of the ego car and important instances, we construct a scene database as shown in Fig.~\ref{fig:databasesql}~(a).
The scene information database includes four tables:
\begin{itemize}
    \item \texttt{Scene Information}: consists of the scene dictionary list. The key for each dictionary is the scene ID, which is the unique identifier of the scene in NuScenes~\cite{caesar2020nuscenes}. The value is a frame ID list, consisting of IDs of frames in the current scene. The details for each frame are referred to the \texttt{Frame Information} table.
     \item \texttt{Frame Information}: consists of the frame dictionary list. The key for each dictionary is the unique ID for a specific frame. The value contains the details for the frame, including the ego-car-information ID and instance-information ID list. The details for the information of the ego car and instances in the frame are referred to the \texttt{Ego Information} and \texttt{Instance Information} tables respectively.
    \item \texttt{Ego Information}: consists of the ego-car information dictionary list. The key for each dictionary is the unique ID for the information of the ego-car in one specific frame. The values include the information: ~\eg, ego car pose, ego car rotation, velocity, road information, camera information, and so on. 
    \item \texttt{Instance Information}: consists of the instance information dictionary list. The key for each dictionary is the unique ID for the information of one instance in one specific frame. The values include the information: ~\eg, instance ID (the unique identifier for one instance,~\eg, a car, across different frames in one scene), instance global and local translations and rotations, velocity, road information, etc. 
\end{itemize}
The relations between different tables in the database are shown in Fig.~\ref{fig:databasesql}~(b) (\textcolor{red}{red dashed arrows}),~\ie, \texttt{Scene Information} and \texttt{Frame Information}, \texttt{Frame Information} and \texttt{Ego Information}, \texttt{Frame Information} and \texttt{Instance Information} are all one-to-many mappings.
The detailed fields and their corresponding types for each table are shown in Table~\ref{tab:database}.
%

\subsection{Task SQLs}~\label{sec:sql}

\vspace{0.5mm}
As shown in Fig.~\ref{fig:databasesql}~(b), we define four types of SQL sets (\ie, perception, prediction, risk, and planning with reasoning), which are used for generating different types of instruction-response pairs.
Each type of task SQL set consists of several subtask SQLs.
For example, the perception SQL set contains six subtask SQLs,~\eg, Closest focuses on finding the objects closest to the ego car, and the other subtasks are similar. 
Note that some high-level SQLs may be inherited from low-level ones following the relational flow of autonomous driving tasks~\cite{hu2023planning}, as shown in \textcolor{blue}{blue dashed arrows} in Fig.~\ref{fig:databasesql}~(b).
For instance, the prediction SQLs are based on the perception ones, the risk SQLs are based on both prediction and perception SQLs, and the planning with reasoning SQL is inherited from prediction and risk ones.

Each subtask SQL consists of a subtask function and an instruction prompt.
We illustrate the detailed algorithms for $17$ subtask SQLs in Algorithm~\ref{alg:distance}-\ref{alg:planning}.
%
In these algorithms, `query($T$, \texttt{*args})' indicates querying the information from the table $T$ with the arguments \texttt{*args}.
After obtaining the results from the task SQLs, we transfer them into the language descriptions by template or GPT-4~\cite{OpenAI_2023}.
Regard the planning with reasoning task as the example, after obtaining the queried results $\{ R, s, m \}$, where $R$, $s$ and $m$ are the risk instance dictionary, the future speed and the future motion of the ego car.
Then, the response is formulated as \texttt{`There are $R$ the ego car. Hence the ego car should be $s$ and move to $m$.'}

\subsection{Evaluation Details}~\label{sec:evaluation}
In this section, we show the computation details for different evaluation metrics,~\ie, MAE, accuracy, MAP, and BLEU.

\paragraph{MAE.} For tasks measured by MAE, we first use the regular expression to obtain the values from the predicted response,~\eg, $\hat{m}$.
Then the MAE is computed by $| \hat{m} - m|$, where $|\cdot|$ indicates the absolute value.

\paragraph{Accuracy.}  There are three kinds of subtasks measured by the accuracy,~\ie, Closest, Status, and Same Road. Since the predictions of the Closest subtask are the instance categories, we formulate the Closest subtask to the classification tasks. Similarly, for the Status subtask, there are totally two different statuses (~\ie, moving and stationary) for the ego car, while other instances have different statuses for different kinds of objects,~\eg, for vehicles, there are moving, stopped and parked; for pedestrians, there are moving, standing and sitting. Hence, the Status subtask can also be regarded as the classification task.
Finally, the responses to the Same Road task are `yes' or `no', which is a binary classification task.

\paragraph{MAP.} We evaluate all subtasks in the risk task by the mean average precision (MAP).
Since risk tasks aim to find the objects that may have a risk influence on the ego car driving.
Hence, we can transfer them to the object detection tasks, which are generally evaluated by MAP.


\paragraph{BLEU.} 
BLEU~\cite{papineni2002bleu} is a classical evaluation metric for caption tasks.
In this paper, we use BLEU~\cite{papineni2002bleu} to evaluate the planning with reasoning task, which is similar to captioning.

\section{More Qualitative Examples}
\begin{figure*}
    \centering    \includegraphics[width=\linewidth,height=0.33\textheight]{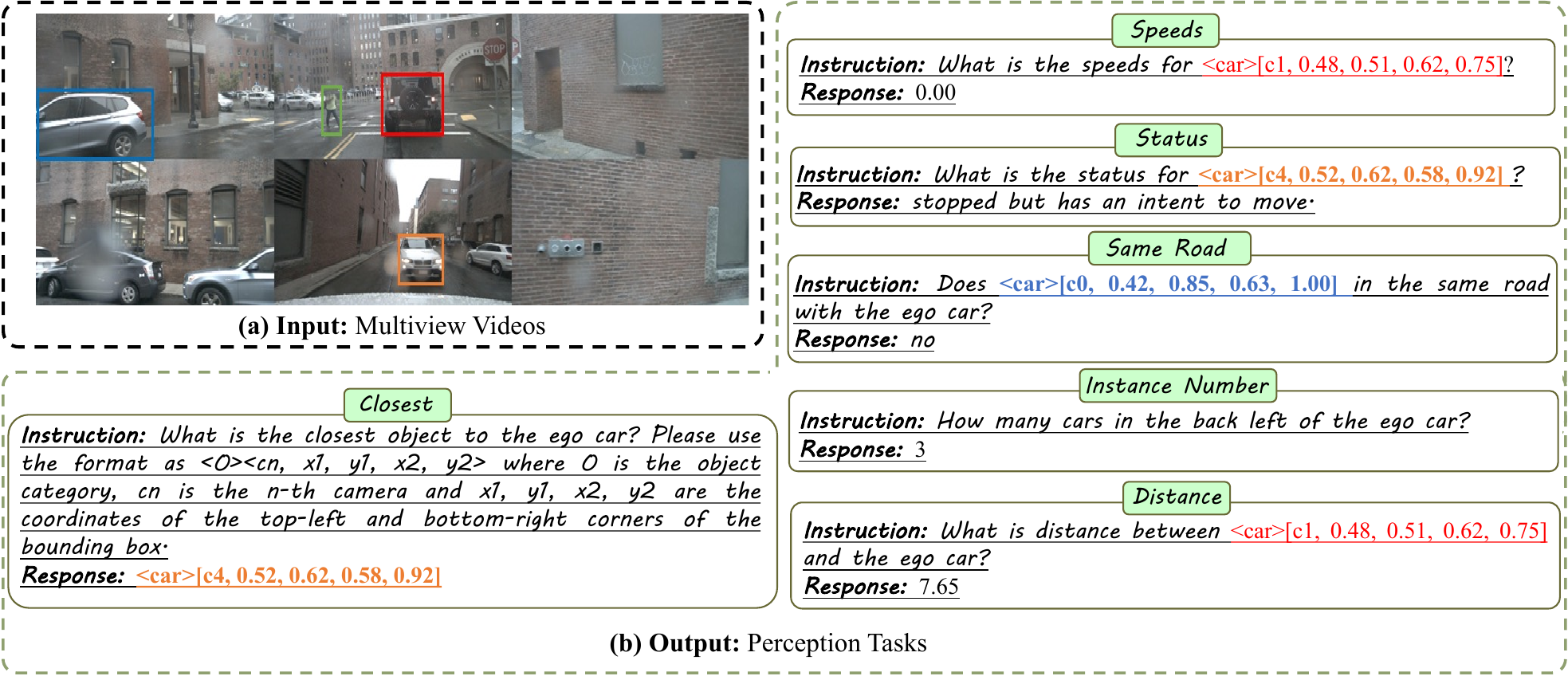}
    \caption{Visualization of our proposed BEV-InMLLM on the perception tasks, which includes six subtasks,~\ie, speeds, status, same road, instance number, distance and closest.
    }
    \label{fig:perception}
\end{figure*}

\begin{figure*}
    \centering
    \includegraphics[width=\linewidth,height=0.33\textheight]{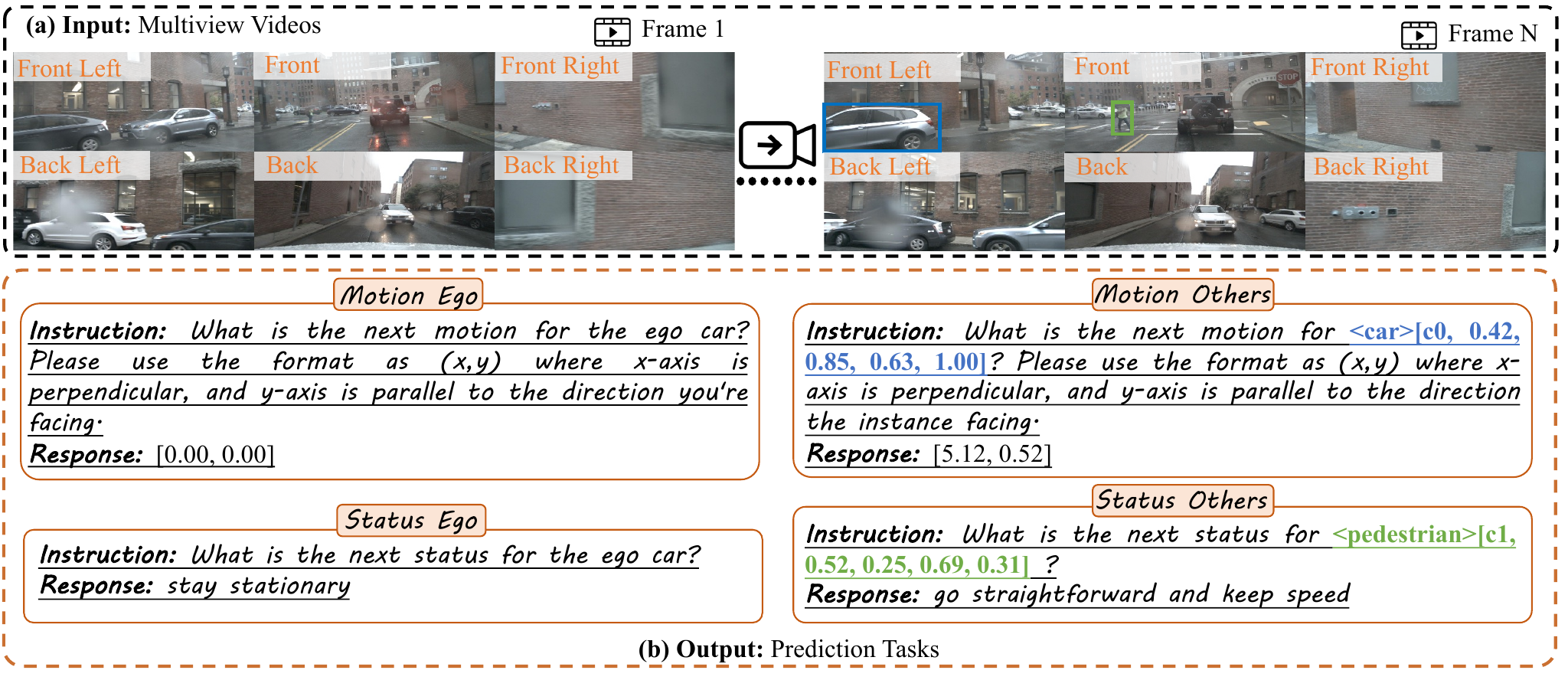}
    \caption{Visualization of our proposed BEV-InMLLM on the prediction tasks, which includes four subtasks,~\ie, motion ego, motion others, status ego and status others.
    }
    \label{fig:prediction}
\end{figure*}

\begin{figure*}
    \centering
    \includegraphics[width=\linewidth,height=0.35\textheight]{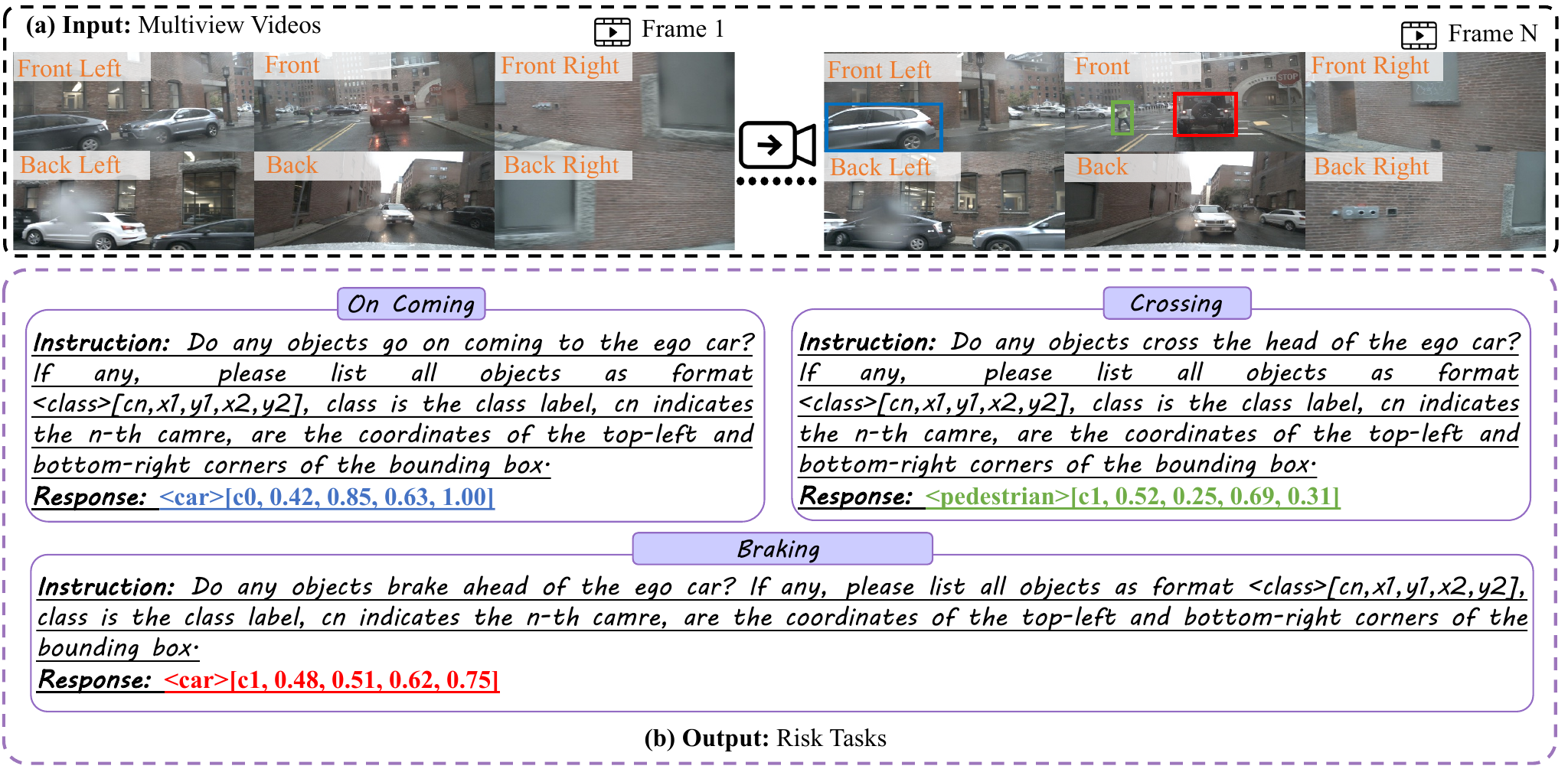}
    \caption{Visualization of our proposed BEV-InMLLM on risk tasks, which includes three subtasks,~\ie, on coming, crossing and braking.
    }
    \label{fig:risk1}
\end{figure*}

\begin{figure*}
    \centering
    \includegraphics[width=\linewidth,height=0.27\textheight]{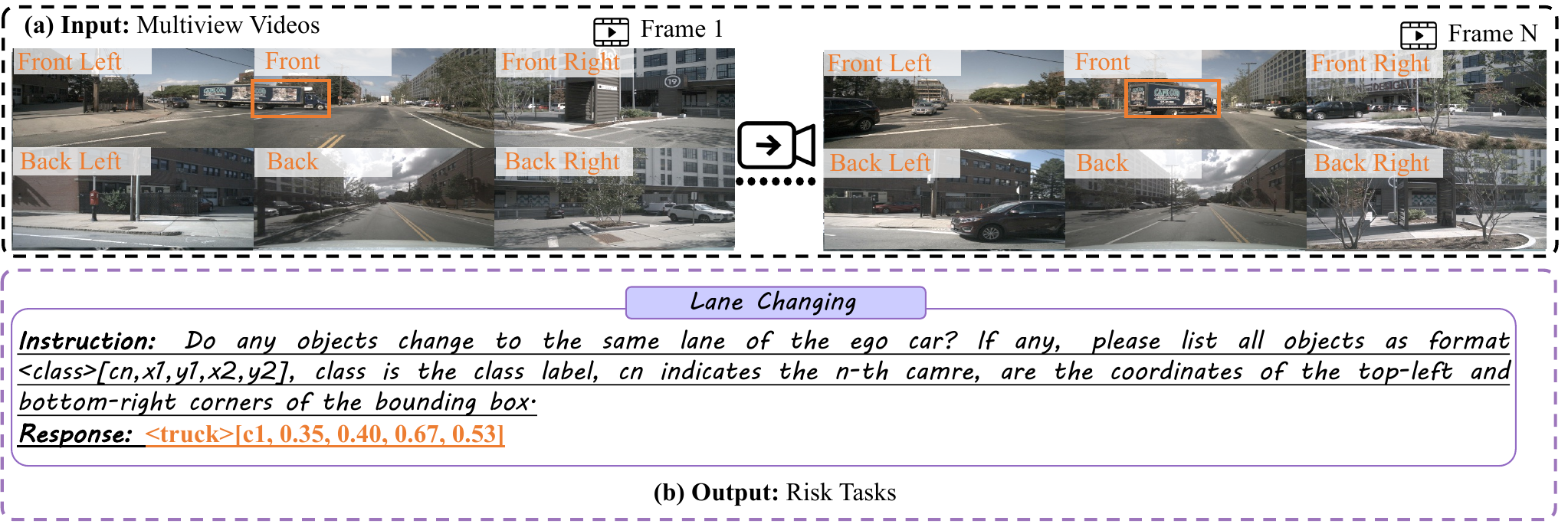}
    \caption{Visualization of our proposed BEV-InMLLM on risk tasks, which includes one subtask,~\ie, lane changing.
    }
    \label{fig:risk2}
\end{figure*}

\begin{figure*}
    \centering
    \includegraphics[width=\linewidth,height=0.33\textheight]{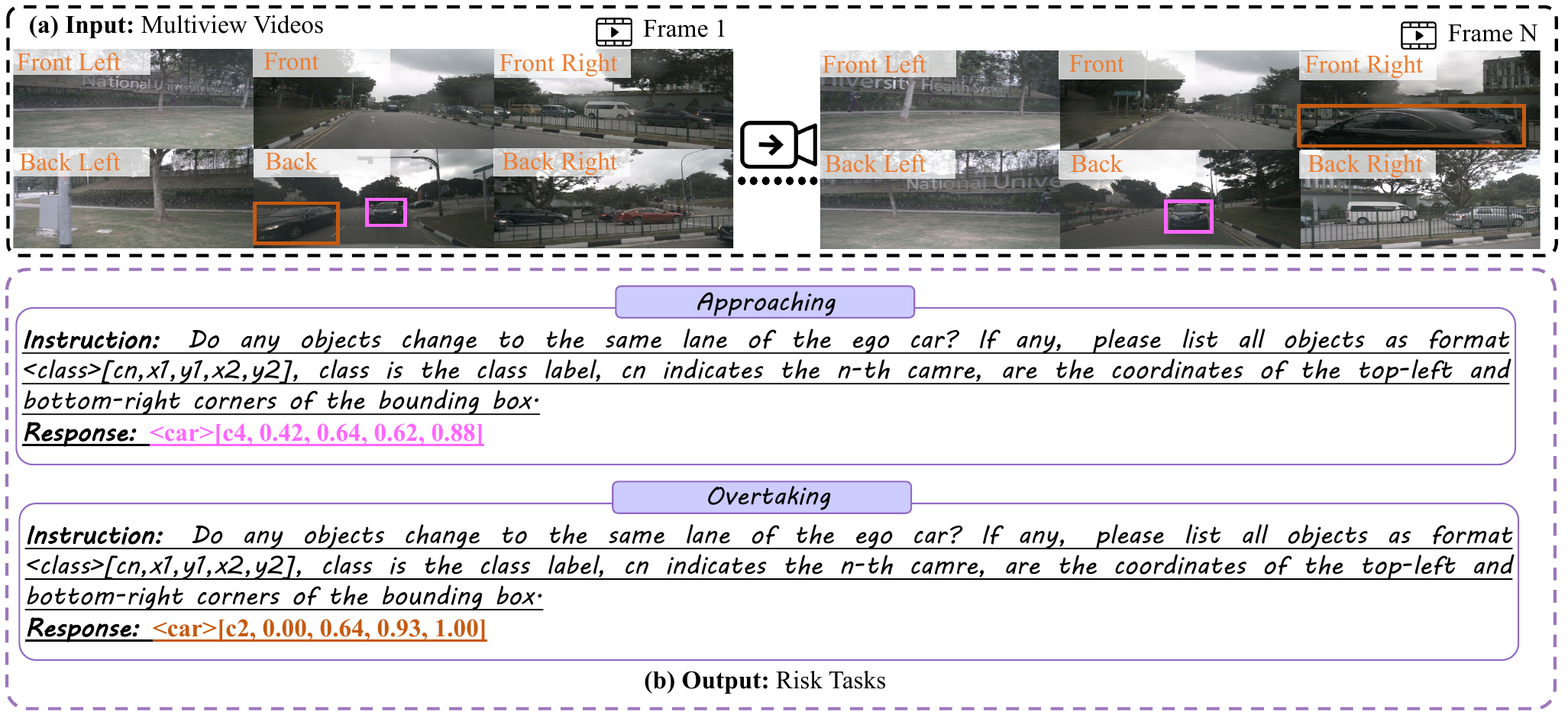}
    \caption{Visualization of our proposed BEV-InMLLM on risk tasks, which includes two subtasks,~\ie, approaching and overtaking.
    }
    \label{fig:risk3}
\end{figure*}

\begin{figure*}
    \centering
    \includegraphics[width=\linewidth,height=0.45\textheight]{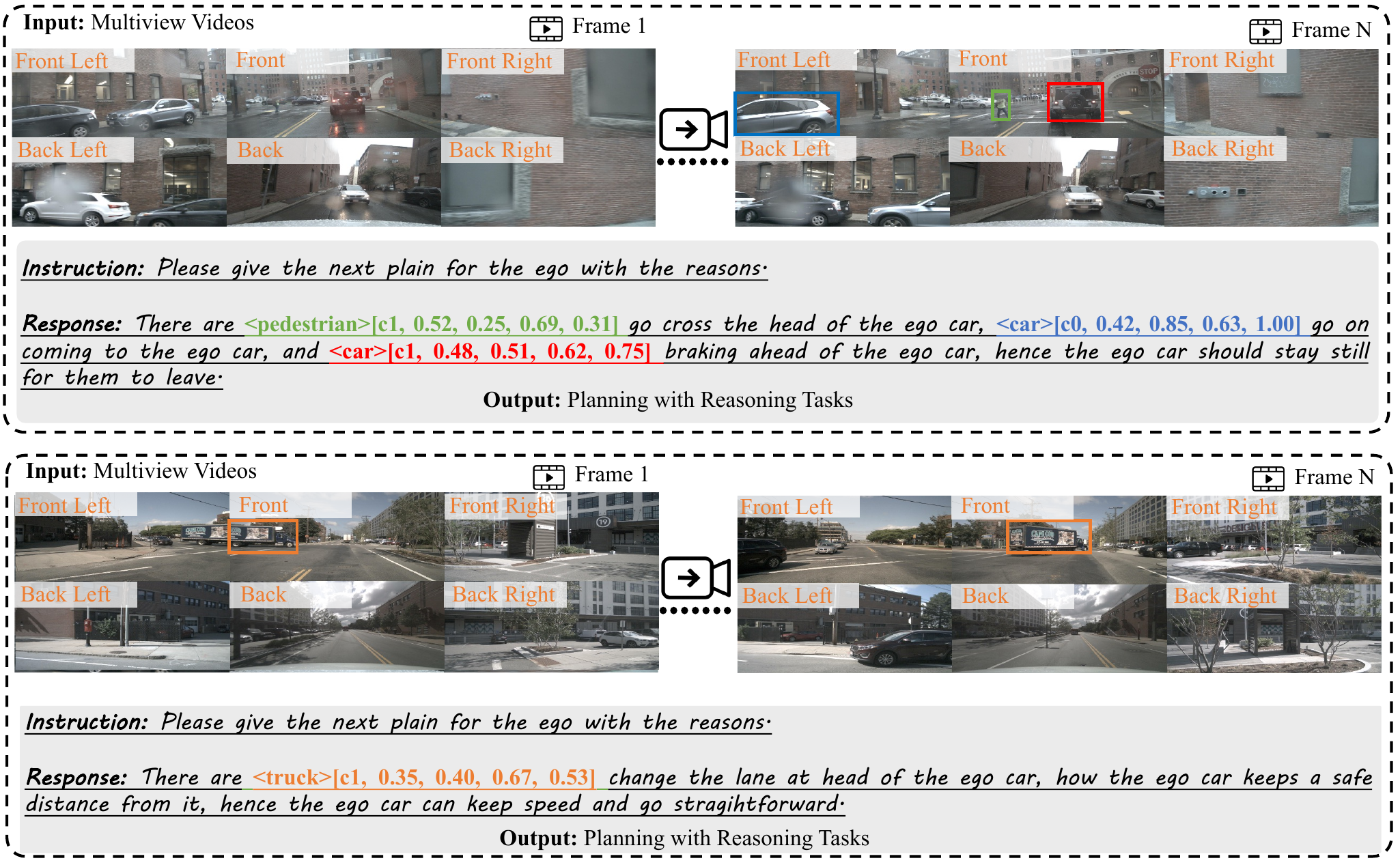}
    \caption{Visualization of our proposed BEV-InMLLM on the planning with reasoning tasks.
    }
    \label{fig:planning}
\end{figure*}

We show more visualization examples for all $17$ subtasks in Fig.~\ref{fig:perception} (perception tasks), Fig.~\ref{fig:prediction} (Prediction tasks), Fig.~\ref{fig:risk1}-\ref{fig:risk3} (risk tasks) and Fig.~\ref{fig:planning} (planning with reasoning tasks).

%

%

%

%

%

%


\begin{algorithm}[t] 
    \centering 
    \small
    \caption{Distance SQL}
    \label{alg:distance} 
    \begin{algorithmic}[1] 
        \STATE\textbf{Input}: Instance information ID: $i$ 
        \STATE\textbf{Instruction prompt $p$}: What is the distance between \texttt{ins} and the ego car?.
        \STATE  $I_i$ = query($T_\text{ins}$, $i$) 
        \STATE  \texttt{ins} = $<$ cn, x1, y1, x2, y2 $>$ = $I_i$[`\textbf{Camera Pos}']
        \STATE  ($x$, $y$) = $I_i$[`\textbf{Local-T}']
        \STATE  $ l = \sqrt{x^2 + y^2}$
        \RETURN $l$
    \end{algorithmic} 
\end{algorithm} 

\begin{algorithm}[t] 
    \centering 
    \small
    \caption{Closest SQL}
    \label{alg:closest} 
    \begin{algorithmic}[1] 
        \STATE\textbf{Input}: Frame ID $i$
        \STATE\textbf{Instruction prompt $p$}: What are the closest objects in \texttt{view} of the ego car?;
        \STATE \texttt{view} = \{{front left, front, front right, back left, back, back right, all} \}.
        \STATE $F_i$ = query($T_\text{frame}$, $i$) 
        \STATE $d_\text{min}$ = inf; $I_\text{min}$ = dict()
        \FOR{$v$ in \texttt{view}}
        \FOR{$n$ in $F_i$[`\textbf{Instance Information ID List}']}  
        \STATE $I_n$ = query($T_\text{ins}$, $n$)
        \IF{ $v$ in $I_n$[`\textbf{Camera Pos}'] or $v$ == all }
        \STATE  ($x$, $y$) = $I_n$[`\textbf{Local-T}']
        \STATE  $ d = \sqrt{x^2 + y^2}$
        \IF{$d_\text{min} < d$}
        \STATE $I_\text{min}[v] = I_n$ 
        \ENDIF
         \ENDIF
        \ENDFOR
        \ENDFOR
        \RETURN $I_\text{min}$
    \end{algorithmic} 
\end{algorithm} 

\begin{algorithm}[t] 
    \centering 
    \small
    \caption{Instance Number SQL}
    \label{alg:instancenumber} 
    \begin{algorithmic}[1] 
        \STATE\textbf{Input}: Frame ID $i$
        \STATE\textbf{Instruction prompt $p$}: How many \texttt{object} in \texttt{view} of the ego car?;
        \STATE \texttt{object} = \{{car, truck, pedestrian, barrier, debris, bicycle, bus, construction, ambulance...} \} 
        \STATE \texttt{view} = \{{front left, front, front right, back left, back, back right, all} \}.
        \STATE $F_i$ = query($T_\text{frame}$, $i$) 
        \STATE  $N$ = dict()
        \FOR{$o$ in \texttt{object}}
        \FOR{$v$ in \texttt{view}}
         \STATE   $N[v][o]=0$
        \FOR{$n$ in $F_i$[`\textbf{Instance Information ID List}']}  
        \STATE $I_n$ = query($T_\text{ins}$, $n$)
        \IF{ $o$ == $I_n$[`\textbf{Category}']}
        \IF{ $v$ in $I_n$[`\textbf{Camera Pos}'] or $v$ == all}
        \STATE  ($x$, $y$) = $I_n$[`\textbf{Local-T}']
        \STATE  $ d = \sqrt{x^2 + y^2}$
        \STATE $N[v][o] ++$ 
        \ENDIF
        \ENDIF
        \ENDFOR
        \ENDFOR
        \ENDFOR
        \RETURN $N$
    \end{algorithmic} 
\end{algorithm} 

\begin{algorithm}[t] 
    \centering 
    \small
    \caption{Speeds SQL}
    \label{alg:speed} 
    \begin{algorithmic}[1] 
        \STATE\textbf{Input}: Instance information ID: $i$ 
        \STATE\textbf{Instruction prompt $p$}: What is the speeds for \texttt{ins}?
        \STATE  $I_i$ = query($T_\text{ins}$, $i$) 
        \STATE  \texttt{ins} = $<$ cn, x1, y1, x2, y2 $>$ = $I_i$[`\textbf{Camera Pos}']
        \STATE  $v$ = $I_i$[`\textbf{Velocity}']
        \RETURN $v$
    \end{algorithmic} 
\end{algorithm} 

\begin{algorithm}[t] 
    \centering 
    \small
    \caption{Status SQL}
    \label{alg:status} 
    \begin{algorithmic}[1] 
        \STATE\textbf{Input}: Instance information ID: $i$ 
        \STATE\textbf{Instruction prompt $p$}: What is the status for \texttt{ins}?
        \STATE  $I_i$ = query($T_\text{ins}$, $i$) 
        \STATE  \texttt{ins} = $<$ cn, x1, y1, x2, y2 $>$ = $I_i$[`\textbf{Camera Pos}']
        \STATE  $s$ = $I_i$[`\textbf{Attribute}']
        \RETURN $s$
    \end{algorithmic} 
\end{algorithm} 

\begin{algorithm}[t] 
    \centering 
    \small
    \caption{SameRoad SQL}
    \label{alg:sameroad} 
    \begin{algorithmic}[1] 
        \STATE\textbf{Input}: Instance information ID: $i$; Frame ID: $n$ 
        \STATE\textbf{Instruction prompt $p$}: Does \texttt{ins} in the same road with the ego car?
        \STATE  $I_i$ = query($T_\text{ins}$, $i$) 
        \STATE  $E_n$ = query($T_\text{ego}$, $n$) 
        \STATE  \texttt{ins} = $<$ cn, x1, y1, x2, y2 $>$ = $I_i$[`\textbf{Camera Pos}']
        \STATE   $r_\text{ins}$ = $I_i$[`\textbf{Road Information}']  
        \STATE  $r_\text{ego}$ = $E_n$[`\textbf{Road Information}']  
        
        \IF{  $r_\text{ins}$ ==$r_\text{ego}$}
             \RETURN yes
        \ELSE
             \RETURN no
        \ENDIF
       
    \end{algorithmic} 
\end{algorithm}

\begin{algorithm}[t] 
    \centering 
    \small
    \caption{Motion Ego SQL}
    \label{alg:motionego} 
    \begin{algorithmic}[1] 
        \STATE\textbf{Input}: Current Frame ID: $i$;  Next Frame ID: $i+1$
        \STATE\textbf{Instruction prompt $p$}: What is the next motion for the ego car?
        \STATE  $E_i$ = query($T_\text{ego}$, $i$); $p_i$ = $E_i$[`\textbf{Pose}']; 
        $r_i$ = $E_i$[`\textbf{Rotation}']
        \STATE $E_{i+1}$ = query($T_\text{ego}$, $i+1$); $p_{i+1}$ = $E_{i+1}$[`\textbf{Pose}'];  
        $r_{i+1}$ = $E_{i+1}$[`\textbf{Rotation}']
        \STATE  $m$ = $r_i^{-1}$.rotate($p_{i+1}$ - $p_{i}$)
        \RETURN $m$
    \end{algorithmic} 
\end{algorithm} 

\begin{algorithm}[t] 
    \centering 
    \small
    \caption{Motion Others SQL}
    \label{alg:motionother} 
    \begin{algorithmic}[1] 
        \STATE\textbf{Input}: Current Frame ID: $i$;  Next Frame ID: $i+1$
        \STATE\textbf{Instruction prompt $p$}: What is the next motion for \texttt{ins}?
        \STATE  $F_i$ = query($T_\text{frame}$, $i$);
        \STATE  $M$ = dict() {\transparent{0.6}\#\# Motion dictionary for instances}
        \FOR{$n$ in $F_i$[`\textbf{Instance Information ID List}']}  
        \STATE $I_n$ = query($T_\text{ins}$, $n$)
        \STATE  \texttt{ins} = $<$ cn, x1, y1, x2, y2 $>$ = $I_n$[`\textbf{Camera Pos}']
        \STATE  $p_n$ = $I_n$[`\textbf{Global-T}'];  $r_n$ = $I_n$[`\textbf{Global-R}']
        \STATE  $d$ = \STATE $d$ = Query($T_\text{ins}$, $I_{n}$[`\textbf{Instance ID}'], $i+1$)
         \STATE  $I_d$ = query($T_\text{ins}$, $d$)
        \STATE  $p_d$ = $I_d$[`\textbf{Global-T}'];  $r_d$ = $I_d$[`\textbf{Global-R}']
        \STATE $M[n]$ = $r_n^{-1}$.rotate($p_{d}$ - $p_{n}$)
        \ENDFOR
        \RETURN $M$
    \end{algorithmic} 
\end{algorithm}

\begin{algorithm}[t] 
    \centering 
    \small
    \caption{Status Ego SQL}
    \label{alg:statusego} 
    \begin{algorithmic}[1] 
        \STATE\textbf{Input}: Current Frame ID: $i$;  Next Frame ID: $i+1$
        \STATE\textbf{Instruction prompt}: What's the next status for the ego car?
        \STATE $m$ = Motion Ego ($i$, $i+1$) {\transparent{0.66}\#\#  Algorithm~\ref{alg:motionego}}
        \STATE  $E_i$ = query($T_\text{ego}$, $i$); $v_i$ = $E_{i}$[`\textbf{Velocity}']
        \STATE $E_{i+1}$ = query($T_\text{ego}$, $i+1$); $v_{i+1}$ = $E_{i+1}$[`\textbf{Velocity}']
        \RETURN $\{v_{i+1}-v_i, m \}$
    \end{algorithmic} 
\end{algorithm}

\begin{algorithm}[t] 
    \centering 
    \small
    \caption{Status Others SQL}
    \label{alg:statusother} 
    \begin{algorithmic}[1] 
        \STATE\textbf{Input}: Current Frame ID: $i$;  Next Frame ID: $i+1$; 
        \STATE\textbf{Instruction prompt}: What's the next status for \texttt{ins}?
        \STATE  $F_i$ = query($T_\text{frame}$, $i$);
        \STATE $M$ = Motion Others($i$,$i+1$) {\transparent{0.66}\#\#  Algorithm~\ref{alg:motionother}}
        \STATE $V$ = dict() {\transparent{0.6}\#\# Speeds dictionary for instances}
        \FOR{$n$ in $F_i$[`\textbf{Instance Information ID List}']}  
        \STATE $I_{n}$ = query($T_\text{ins}$, $n$)
        \STATE $d$ = $I_{n}$[`\textbf{Instance ID}']
        \STATE $v_i$ = Speeds($n$); $v_{i+1}$ = Speeds($d$) {\transparent{0.66}\#\# Algorithm~\ref{alg:speed}}
        \STATE $V[n]=v_{i+1}-v_i$    
        \ENDFOR
        \RETURN $\{V, M\}$
    \end{algorithmic} 
\end{algorithm}

\begin{algorithm}[t] 
    \centering 
    \small
    \caption{Overtaking SQL}
    \label{alg:overtaking} 
    \begin{algorithmic}[1] 
        \STATE\textbf{Input}: Previous Frame ID: $i-1$; Current Frame ID: $i$;  Next Frame ID: $i+1$; Threshold distance to the ego car: $dis$
        \STATE\textbf{Instruction prompt}: Do any objects overtake the ego car?
        \STATE  $F_i$ = query($T_\text{frame}$, $i$);
        \STATE $M_{i-1}$ = Motion Others($i$, $i-1$) {\transparent{0.66}\#\# Algorithm~\ref{alg:motionother}}
        \STATE $M_i$ = Motion Others($i$, $i+1$) {\transparent{0.66}\#\# Algorithm~\ref{alg:motionother}}
        \STATE $O$ = list() {\transparent{0.6}\#\# Instance list}
        \FOR{$n$ in $F_i$[`\textbf{Instance Information ID List}']}  
        \STATE $I_{n}$ = query($T_\text{ins}$, $n$)
        \STATE $d$ = Query($T_\text{ins}$, $I_{n}$[`\textbf{Instance ID}'], $i-1$)
        \STATE $v_i$ = Speeds($n$); $v_{i-1}$ = Speeds($d$) {\transparent{0.66}\#\# Algorithm~\ref{alg:speed}}
        \IF{$M_{i-1}[d][0]<0$ and $M_{i-1}[d][0]>0$ and $M_{i-1}[d][1]<dis$ and $M_{i-1}[d][1]<dis$ and $v_i>0$ and $v_{i-1}>0$}
            \STATE $O$.append($I_{n}$)
        \ENDIF
        \ENDFOR
        \RETURN $O$
    \end{algorithmic} 
\end{algorithm} 

\begin{algorithm}[t] 
    \centering 
    \small
    \caption{On Coming SQL}
    \label{alg:oncoming} 
    \begin{algorithmic}[1] 
        \STATE\textbf{Input}: Previous Frame ID: $i-1$; Current Frame ID: $i$;  Next Frame ID: $i+1$; Threshold distance to the ego car: $dis$
        \STATE\textbf{Instruction prompt}: Do any objects go on coming to the ego car?
        \STATE  $F_i$ = query($T_\text{frame}$, $i$);
        \STATE $M_{i-1}$ = Motion Others($i$, $i-1$) {\transparent{0.66}\#\#  Algorithm~\ref{alg:motionother}}
        \STATE $M_i$ = Motion Others($i$, $i+1$) {\transparent{0.6}\#\#  Algorithm~\ref{alg:motionother}}
        \STATE $O$ = list() {\transparent{0.6}\#\# Instance list}
        \FOR{$n$ in $F_i$[`\textbf{Instance Information ID List}']}  
        \STATE $I_{n}$ = query($T_\text{ins}$, $n$)
        \STATE $d$ = Query($T_\text{ins}$, $I_{n}$[`\textbf{Instance ID}'], $i-1$)
        \STATE $v_i$ = Speeds($n$); $v_{i-1}$ = Speeds($d$) {\transparent{0.66}\#\# Algorithm~\ref{alg:speed}}
        \IF{$M_{i-1}[d][0]>0$ and $M_{i-1}[d][0]>0$ and $M_{i}[d][0] < M_{i-1}[d][0]$ and $M_{i-1}[d][1]<dis$ and $M_{i}[d][1]<dis$ and $abs(M_{i}[d][1]-M_{i-1}[d][1])<dis)$and $v_i>0$ and $v_{i-1}>0$}
        \STATE $O$.append($I_{n}$)
        \ENDIF
        \ENDFOR
        \RETURN $O$
    \end{algorithmic} 
\end{algorithm} 

\begin{algorithm}[t] 
    \centering 
    \small
    \caption{Approaching SQL}
    \label{alg:approaching} 
    \begin{algorithmic}[1] 
        \STATE\textbf{Input}: Previous Frame ID: $i-1$; Current Frame ID: $i$;  Next Frame ID: $i+1$; Threshold distance to the ego car: $dis$
        \STATE\textbf{Instruction prompt}: Do any objects approach the ego car?
        \STATE  $F_i$ = query($T_\text{frame}$, $i$);
        \STATE $M_{i-1}$ = Motion Others($i$, $i-1$) {\transparent{0.66}\#\# Algorithm~\ref{alg:motionother}}
        \STATE $M_i$ = Motion Others($i$, $i+1$) {\transparent{0.66}\#\# Algorithm~\ref{alg:motionother}}
        \STATE $O$ = list()  {\transparent{0.6}\#\# Instance list}
        \FOR{$n$ in $F_i$[`\textbf{Instance Information ID List}']}  
        \STATE $I_{n}$ = query($T_\text{ins}$, $n$)
        \STATE $d$ = Query($T_\text{ins}$, $I_{n}$[`\textbf{Instance ID}'], $i-1$)
        \STATE $v_i$ = Speeds($n$); $v_{i-1}$ = Speeds($d$) {\transparent{0.66}\#\# Algorithm~\ref{alg:speed}}
        \IF{ $M_{i}[d][0] < M_{i-1}[d][0]$ and $M_{i-1}[d][1]<dis$ and $M_{i}[d][1]<dis$ and $abs(M_{i}[d][1]-M_{i-1}[d][1])<dis)$and $v_i>0$ and $v_{i-1}>0$}
        \STATE $O$.append($I_{n}$)
        \ENDIF
        \ENDFOR
        \RETURN $O$
    \end{algorithmic} 
\end{algorithm} 

\begin{algorithm}[t] 
    \centering 
    \small
    \caption{Crossing SQL}
    \label{alg:crossing} 
    \begin{algorithmic}[1] 
        \STATE\textbf{Input}: Previous Frame ID: $i-1$; Current Frame ID: $i$;  Next Frame ID: $i+1$; Threshold distance to the ego car: $dis$; Threshold distance for $x$ direction: $dis_x$; Threshold distance for $y$ direction: $dis_y$
        \STATE\textbf{Instruction prompt}: Do any objects cross the head of the ego car?
        \STATE  $F_i$ = query($T_\text{frame}$, $i$);
        \STATE $M_{i-1}$ = Motion Others($i$, $i-1$) {\transparent{0.66}\#\# Algorithm~\ref{alg:motionother}}
        \STATE $M_i$ = Motion Others($i$, $i+1$) {\transparent{0.66}\#\# Algorithm~\ref{alg:motionother}}
        \STATE $O$ = list() {\transparent{0.6}\#\# Instance list}
        \FOR{$n$ in $F_i$[`\textbf{Instance Information ID List}']}  
        \STATE $I_{n}$ = query($T_\text{ins}$, $n$)
        \STATE $d$ = Query($T_\text{ins}$, $I_{n}$[`\textbf{Instance ID}'], $i-1$)
         \STATE $l_i$ = Distance($n$); $l_{i-1}$ = Distance($d$) {\transparent{0.66}\#\# Algorithm~\ref{alg:distance}}
        \STATE $v_i$ = Speeds($n$); $v_{i-1}$ = Speeds($d$) {\transparent{0.66}\#\# Algorithm~\ref{alg:speed}}
        \IF{ $l_i<dis$ and $l_{i-1}<dis$ and $abs(M_{i}[d][0]-M_{i-1}[d][0])<dis_x)$ and $abs(M_{i}[d][1]-M_{i-1}[d][1])>dis_y)$ and $v_i>0$ and $v_{i-1}>0$}
        \STATE $O$.append($I_{n}$)
        \ENDIF
        \ENDFOR
        \RETURN $O$
    \end{algorithmic} 
\end{algorithm} 

\begin{algorithm}[t] 
    \centering 
    \small
    \caption{Braking SQL}
    \label{alg:braking} 
    \begin{algorithmic}[1] 
        \STATE\textbf{Input}: Previous Frame ID: $i-1$; Current Frame ID: $i$;  Next Frame ID: $i+1$; Threshold distance to the ego car: $dis$; Threshold distance for $x$ direction: $dis_x$; Threshold distance for $y$ direction: $dis_y$; Threshold speed: $s$
        \STATE\textbf{Instruction prompt}: Do any objects brake ahead of the ego car?
        \STATE  $F_i$ = query($T_\text{frame}$, $i$);
        \STATE $M_{i-1}$ = Motion Others($i$, $i-1$) {\transparent{0.66}\#\# Algorithm~\ref{alg:motionother}}
        \STATE $M_i$ = Motion Others($i$, $i+1$) {\transparent{0.66}\#\# Algorithm~\ref{alg:motionother}}
        \STATE $O$ = list() {\transparent{0.6}\#\# Instance list}
        \FOR{$n$ in $F_i$[`\textbf{Instance Information ID List}']}  
        \STATE $I_{n}$ = query($T_\text{ins}$, $n$)
        \STATE $d$ = Query($T_\text{ins}$, $I_{n}$[`\textbf{Instance ID}'], $i-1$)
         \STATE $l_i$ = Distance($n$); $l_{i-1}$ = Distance($d$) {\transparent{0.66}\#\# Algorithm~\ref{alg:distance}}
        \STATE $v_i$ = Speeds($n$); $v_{i-1}$ = Speeds($d$) {\transparent{0.66}\#\# Algorithm~\ref{alg:speed}}
        \IF{ $l_i< l_{i-1} < dis$  and $abs(M_{i}[d][0]-M_{i-1}[d][0])>dis_x)$ and $M_{i}[d][1]< dis_y$ and $M_{i-1}[d][1])<dis_y)$ and $v_{i-1}>s$ and $v_i < s$}
        \STATE $O$.append($I_{n}$)
        \ENDIF
        \ENDFOR
        \RETURN $O$
    \end{algorithmic} 
\end{algorithm} 

\begin{algorithm}[t] 
    \centering 
    \small
    \caption{Lane Chaning SQL}
    \label{alg:lanechaning} 
    \begin{algorithmic}[1] 
        \STATE\textbf{Input}: Previous Frame ID: $i-1$; Current Frame ID: $i$;  Next Frame ID: $i+1$; Threshold distance to the ego car: $dis$; Threshold speed: $s$
        \STATE\textbf{Instruction prompt}: Do any objects change to the same lane of the ego car?
        \STATE  $F_i$ = query($T_\text{frame}$, $i$);
        \STATE $M_{i-1}$ = Motion Others($i$, $i-1$) {\transparent{0.66}\#\# Algorithm~\ref{alg:motionother}}
        \STATE $M_i$ = Motion Others($i$, $i+1$) {\transparent{0.66}\#\# Algorithm~\ref{alg:motionother}}
        \STATE $O$ = list() {\transparent{0.6}\#\# Instance list}
        \FOR{$n$ in $F_i$[`\textbf{Instance Information ID List}']}  
        \STATE $I_{n}$ = query($T_\text{ins}$, $n$)
        \STATE $d$ = Query($T_\text{ins}$, $I_{n}$[`\textbf{Instance ID}'], $i-1$)
         \STATE $l_i$ = Distance($n$); $l_{i-1}$ = Distance($d$) {\transparent{0.66}\#\# Algorithm~\ref{alg:distance}}
        \STATE $v_i$ = Speeds($n$); $v_{i-1}$ = Speeds($d$) {\transparent{0.66}\#\# Algorithm~\ref{alg:speed}}
        \STATE $r_i$ = SameRoad($n$); $r_{i-1}$ = SameRoad($d$) {\transparent{0.66}\#\# Algorithm~\ref{alg:sameroad}}
        \IF{ $l_i< l_{i-1} < dis$ and $r_i$==yes and $r_{i-1}$==no and $v_{i-1}>s$ and $v_i>s$}
        \STATE $O$.append($I_{n}$)
        \ENDIF
        \ENDFOR
        \RETURN $O$
    \end{algorithmic} 
\end{algorithm}

\begin{algorithm}[t] 
    \centering 
    \small
    \caption{Planning with Reasoning SQL}
    \label{alg:planning} 
    \begin{algorithmic}[1] 
        \STATE\textbf{Input}: Previous Frame ID: $i-1$; Current Frame ID: $i$;  Next Frame ID: $i+1$; Risk dictionary: \texttt{risks} = \{ `Overking', `On Coming', `Approaching', `Crossing', `Braking', `Lane Changing' \}
        \STATE\textbf{Instruction prompt}: Please give the next plan for the ego car with reasons.
        \STATE $R$ = dict() {\transparent{0.6}\#\# Risk instance dictionary}
        \STATE $m$ = Motion Ego($i$, $i+1$) {\transparent{0.66}\#\# Algorithm~\ref{alg:motionego}}
        \STATE $s$ = Status Ego($i$, $i+1$) {\transparent{0.66}\#\# Algorithm~\ref{alg:statusego}}
        \FOR{\texttt{risk} in \texttt{risks}}
        \STATE $R[\texttt{risk}]$ = \texttt{risk}($i-1$, $i$, $i+1$) {\transparent{0.66}\#\# Algorithm~\ref{alg:overtaking}-\ref{alg:lanechaning}}
        \ENDFOR
        \RETURN $\{ R, S, M \}$
    \end{algorithmic} 
\end{algorithm}


\end{document}